\documentclass{article} % For LaTeX2e
\usepackage{iclr2025_conference,times}

\usepackage{amsmath,amsfonts,bm}

\def\eqref#1{equation~\ref{#1}}
\def\1{\bm{1}}

\DeclareMathAlphabet{\mathsfit}{\encodingdefault}{\sfdefault}{m}{sl}
\SetMathAlphabet{\mathsfit}{bold}{\encodingdefault}{\sfdefault}{bx}{n}

\usepackage{hyperref}
\usepackage{url}
\usepackage{array}  % For advanced table features
\usepackage{boldline}
\usepackage{booktabs}  % For professional-looking tables
\usepackage{xcolor}    % For custom colors
\usepackage{colortbl}  % For coloring table cells
\usepackage{siunitx}   % For aligning numbers
\usepackage{makecell}  % For line breaks in cells
\usepackage{caption}   % For customizing captions
\usepackage{flushend}
\usepackage{subcaption}
\usepackage{amsmath}
\usepackage{graphicx,float}
\usepackage{soul} % for strike-through \st command
\usepackage{tabularray}
\usepackage{wrapfig}
\usepackage{multirow}
\usepackage{mdframed}
\usepackage{tcolorbox}
\usepackage{fancybox}

\newcommand{\std}[1]{\textcolor{black}{\scriptsize{$\pm #1$}}}

\usepackage{amsmath,amsfonts,amssymb}       % blackboard math symbols
\usepackage{nicefrac}       % compact symbols for 1/2, etc.
\usepackage{microtype}      % microtypography
\usepackage{xcolor}         % colors
\usepackage{float}
\usepackage{graphicx}

\usepackage{booktabs} % For better table lines
\usepackage{array}    % For table wrapping and better column spacing

\usepackage{algorithm,algorithmic}

\usepackage{chngcntr}

\usepackage{enumitem}

\usepackage{amssymb}
\usepackage{mathtools}
\usepackage{amsthm}
\usepackage{textcomp}

\usepackage[createShortEnv,conf={no link to proof, text link},commandRef=autoref]{proof-at-the-end}

\definecolor{lightgray}{gray}{0.9}
\definecolor{midgray}{gray}{0.7}

\iclrfinalcopy

\title{Mitigating Shortcut Learning with Diffusion Counterfactuals and Diverse Ensembles}

\author{Luca Scimeca \\%\thanks{ Use footnote for providing further information
Mila - Quebec AI Institute and Université de Montréal, Quebec\\
\texttt{luca.scimeca@mila.quebec} \\
\AND
Alexander Rubinstein, Seong Joon Oh \\
Eberhard-Karls-Universität Tübingen, Germany \\
\AND
Damien Teney \\
Idiap Research Institute, Switzerland \\
\AND
Yoshua Bengio \\%\thanks{ Use footnote for providing further information
Mila - Quebec AI Institute and Université de Montréal, Quebec\\
CIFAR Senior Fellow\\
}

\begin{document}

\maketitle

\vspace{-20pt}
\begin{abstract}

% placeholder abstract -- to be changed/modified
  Spurious correlations in the data, where multiple cues are predictive of the target labels, often lead to a phenomenon known as shortcut learning, where a model relies on erroneous, easy-to-learn cues while ignoring reliable ones. In this work, we propose \emph{DiffDiv} an ensemble diversification framework exploiting Diffusion Probabilistic Models (DPMs) to mitigate this form of bias. We show that at particular training intervals, DPMs can generate images with novel feature combinations, even when trained on samples displaying correlated input features. We leverage this crucial property to generate synthetic counterfactuals to increase model diversity via ensemble disagreement. We show that DPM-guided diversification is sufficient to remove dependence on shortcut cues, without a need for additional supervised signals. We further empirically quantify its efficacy on several diversification objectives, and finally show improved generalization and diversification on par with prior work that relies on auxiliary data collection.
\end{abstract}

\section{Introduction}
\label{sec:intro}

% PROBLEM STATEMENT
% ---- What is shortcut learning
Deep Neural Networks (DNNs) have achieved unparalleled success in countless tasks across diverse domains. However, they are not devoid of pitfalls. 
One such downside manifests in the form of shortcut learning, a phenomenon whereby models latch onto simple, non-essential cues that are spuriously correlated with target labels in the training data~\citep{Geirhos2020, shah2020pitfalls, scimeca2022shortcut}. Often engendered by the under-specification present in data, this simplicity bias presents easy to learn \emph{shortcuts} that allow accurate prediction at training time, irrespective of a model's alignment to the downstream task. For instance, previous work has found models to incorrectly rely on background signals for object recognition~\citep{xiao2021noise, Beery_2018_ECCV}, or to rely on non-clinically relevant metal tokens to predict patient conditions from X-Ray images~\citep{Zech2018-qx}.
% --- why is it a problem
Shortcut learning has often been found to lead to significant drops in generalization performance~\citep{AgrawaL_2018_CVPR, UnbiasedLook2011, ShortcutRemoval2020, Repair2019, Zech2018-qx}.
Leveraging shortcut cues can also be harmful when deploying models in sensitive settings, for example, in the reinforcement of harmful biases over the use of sensitive attributes such as gender or skin color~\citep{EstimatingAndMitigating2019, InvestigatingBias2020, scimeca2022shortcut}.

Addressing simplicity biases in machine learning has been a focus of extensive research, often aimed to encourage models to use a more diversified set of predictive cues. A variety of these methods have been input-centric, designed to drive models to focus on different areas of the input space~\citep{teney2022predicting, nicolicioiu2023learning}, while others have focused on diversification strategies that rely on auxiliary data for \emph{prediction  
disagreement}~\citep{pagliardini2022agree, lee2022diversify}. The latter approaches, in particular, have been instrumental in developing functionally diverse models that exhibit robustness to shortcut biases. However, they are limited by their required access to auxiliary data that is often challenging to obtain.

% GOAL/constraints
The primary objective of this work is to mitigate shortcut learning tendencies, particularly when they result in strong, unwarranted biases, access to \emph{ood} data is expensive, and different features may rely on similar areas of the input space. To achieve this objective, we propose \emph{DiffDiv}, an ensemble framework relying on unlabelled \emph{ood} data for shortcut mitigation by ensemble diversification. To overcome the challenges of the past, we aim to synthetically generate the data for model diversification, and thus avoid the impracticality of \emph{ood} data collection. We posit that the synthetic data should: first, lie in the manifold of the data of interest; and second, be at least partially free of the same shortcuts as the original training data. We leverage Diffusion Probabilistic Models (DPMs) to generate synthetic data for ensemble disagreement, leading to strong model diversification.

% plan 
We observe, and empirically show, that even in the presence of correlated features in the data, DPMs can be used to generate synthetic counterfactuals that break the shortcut signals present at training time. We further show that shortcut mitigation can be achieved via ensemble disagreement on these DPM-generated samples. Remarkably, our experiments confirm that the extent and quality of our diffusion-guided ensemble diversification is on par with existing methods that rely on additional data. 

Our contributions are the following:
\begin{enumerate}
    \item We show that DPMs can generate feature compositions beyond data exhibiting correlated input features.
    \item We demonstrate that ensemble disagreement is sufficient for shortcut cue mitigation.
    \item We propose \emph{DiffDiv}, a framework to achieve bias mitigation through diversification based on diffusion counterfactuals.
    \item We show appropriately trained DPM counterfactuals can lead to state-of-the-art diversification and shortcut bias mitigation, without the need for additional data collection.
\end{enumerate}
Moreover, we apply for the first time the Wisconsin Card Sorting Test for Machine Learners (WCST-ML)~\citep{scimeca2022shortcut} to the CelebA face dataset, exposing biased inference for \emph{pale skin} features. We provide an overview of related work in Suppl. \autoref{sup:sec:related_work}

\begin{figure}
\centering
\vspace{-20pt}
\includegraphics[width=.9\linewidth]{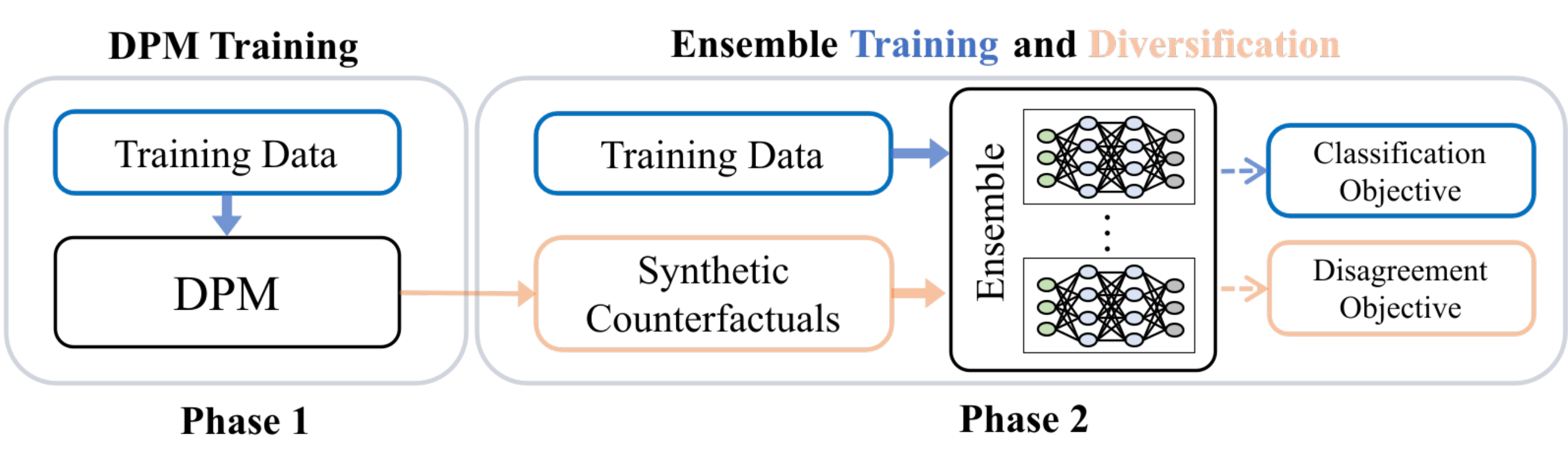}
\caption{DiffDiv: We sample from a DPM to generate synthetic counterfactuals showcasing emergent novel feature combinations. These samples are then utilized to build a diverse model ensemble via different ensemble disagreement objectives.} 
\label{fig:diffdiv}
\vspace{-5pt}
\end{figure}

% -------------------------------------------------------------
% -------------------------------------------------------------
% -------------------------------------------------------------
% ---------------------   METHODS    -----------------------
% -------------------------------------------------------------
% -------------------------------------------------------------
% -------------------------------------------------------------
% -------------------------------------------------------------

\section{Methods}
\label{sec:methods}

\subsection{DiffDiv Overview} \label{sec:overview}
We apply DiffDiv in two stages. The first stage corresponds to training a DPM on the dataset of interest. The second stage corresponds to training an ensemble on the same data and jointly optimize two objectives, a standard classification objective, and a diversification objective. The standard classification objective is performed on real data, while the diversification objective is performed on synthetic counterfactuals generated by the pre-trained DPM. We only consider a fixed small set of 3k counterfactuals in all experiments, and show it to be sufficient for bias mitigation, allowing our method to scale well in complex settings. Please see Suppl. \S \autoref{sup:sec:res_diversification_all_ood} for ablations on counterfactual set size, and their (negligible) impact on \emph{DiffDiv}'s performance.

\subsection{DPMs and Efficient Sampling} \label{sec:diff_training}
% Based on https://arxiv.org/pdf/2202.00512.pdf

We utilize Diffusion Probabilistic Models (DPMs) to generate synthetic data for our experiments. DPMs operate by iteratively adding or removing noise from an initial data point \( x \) through a stochastic process governed by a predefined noise schedule. Let $\mathbf{z}=\left\{\mathbf{z}_t \mid t \in[0,1]\right\}$ be a latent variable conditioned on $t$, and characterized by a noise-to-signal ratio $\lambda_t=\log \left[\alpha_t^2 / \sigma_t^2\right]$ decreasing monotonically with $t$. In the forward process, noise is added to $x$ to transform it into $z_t$ \citep{salimans2022progressive}:

\begin{equation}
q\left(\mathbf{z}_t \mid \mathbf{x}\right)=\mathcal{N}\left(\mathbf{z}_t ; \alpha_t \mathbf{x}, \sigma_t^2 \mathbf{I}\right), \quad q\left(\mathbf{z}_t \mid \mathbf{z}_s\right)=\mathcal{N}\left(\mathbf{z}_t ;\left(\alpha_t / \alpha_s\right) \mathbf{z}_s, \sigma_{t \mid s}^2 \mathbf{I}\right)
\end{equation}
where and $0 \leq s<t \leq 1$, and  $\sigma_{t \mid s}^2=\left(1-e^{\lambda_t-\lambda_s}\right) \sigma_t^2$. 
We let $\alpha_t$ follow a cosine schedule, thus $\alpha_t=\cos (0.5 \pi t)$.

The reverse process then aims to reconstruct $x = z_{0}$ by iteratively denoising $z_t$ into $z_{s}$, starting from $z_1 \sim \mathcal{N}(0, I)$. To facilitate efficient sampling, we employ Denoising Diffusion Implicit Models (DDIM)~\citep{song2020denoising}, a first-order ODE solver for DPMs~\citep{salimans2022progressive, lu2022dpm}, utilizing a predictor-corrector scheme to minimize the number of sampling steps:

\begin{equation}
\begin{aligned}
\mathbf{z}_s & =\alpha_s \hat{\mathbf{x}}_\theta\left(\mathbf{z}_t\right)+\sigma_s \frac{\mathbf{z}_t-\alpha_t \hat{\mathbf{x}}_\theta\left(\mathbf{z}_t\right)}{\sigma_t} \\
& =e^{\left(\lambda_t-\lambda_s\right) / 2}\left(\alpha_s / \alpha_t\right) \mathbf{z}_t+\left(1-e^{\left(\lambda_t-\lambda_s\right) / 2}\right) \alpha_s \hat{\mathbf{x}}_\theta\left(\mathbf{z}_t\right)
\end{aligned}
\end{equation}

The term $\hat{\mathbf{x}}_\theta\left(\mathbf{z}_t\right)$ is the neural network's output, predicting the denoised data from the noisy observation at timestep $t$. And train the denoising model by: 

\begin{align} \label{eq:diff_objective}
    % L_\theta = \frac{1 + \alpha_t^2}{\sigma_t^2} \| x - \hat{x}_\theta(x_t) \|^2,
    L_\theta=\left(1+\frac{\alpha_t^2}{\sigma^2}\right)\left\|\mathbf{x}-\hat{\mathbf{x}}_t\right\|_2^2
\end{align}

We consider DPMs trained at different fidelity levels, and use the `number of diffusion training epochs' as a proxy for the DPM fidelity on the modeled distribution.

\subsection{Ensemble Training and Diversification} \label{sec:ensemble_training}

In our setup, we wish to train a set of models within an ensemble while encouraging model diversity. Let $f_i$ denote the i$^{th}$ model predictions within an ensemble consisting of $N_m$ models. Each model is trained on a joint objective comprising the conventional cross-entropy loss with the target labels, complemented by a diversification term computed on synthetic counterfactuals, represented as ${L}_{\text{div}}$.
\begin{equation}
\mathcal{L} = \mathcal{L}_{\text{xent}} + \gamma\ \mathcal{L}^{\text{obj}}_{\text{div}}
\end{equation} 
where $\mathcal{L}_{\text{xent}}$ is the cross-entropy loss, $\mathcal{L}^{\text{obj}}_{\text{div}}$ is the diversification term for a particular objective, and $\gamma$ is a hyper-parameter used to modulate the importance of diversification within the optimization objective. To impart diversity to the ensemble, we investigated five diversification objectives denoted by ${{obj} \in \{\ \text{div},\ \text{cross},\ L_1,\ L_2,\ \text{kl}\}}$. 

% L_1 L_2 OBJECTIVES

The $L_1$ and $L_2$ baseline objectives were designed to induce diversity by maximizing the distance between each model output and the moving average of the ensemble prediction, thus:

% \setlength{\abovedisplayskip}{5pt}
% \begin{align}
%     L^{L_1}_{\text{reg}} &= -\frac{1}{N_m} \sum_{i=1}^{N_m} \left\| f_i - \frac{1}{N_m} \sum_{j=1}^{N_m} f_j \right\|_1 
% \end{align}
% \begin{align}
%     L^{L_2}_{\text{reg}} &= -\frac{1}{N_m} \sum_{i=1}^{N_m} \left\| f_i - \frac{1}{N_m} \sum_{j=1}^{N_m} f_j \right\|_2^2
% \end{align}

\begin{minipage}{\linewidth}
    \hspace{20pt}
    \begin{minipage}{0.495\linewidth}
        \begin{equation*}
            L^{L_1}_{\text{reg}} = -\frac{1}{N_m} \sum_{i=1}^{N_m} \left\| f_i - \frac{1}{N_m} \sum_{j=1}^{N_m} f_j \right\|_1
            \label{eq:L1}
        \end{equation*}
    \end{minipage}
    \hspace{-40pt}
    \begin{minipage}{0.495\linewidth}
        \begin{equation*}
            L^{L_2}_{\text{reg}} = -\frac{1}{N_m} \sum_{i=1}^{N_m} \left\| f_i - \frac{1}{N_m} \sum_{j=1}^{N_m} f_j \right\|_2^2
            \label{eq:L2}
        \end{equation*}
    \end{minipage}
\end{minipage}

% The strengths of these components are controlled by hyperparameters $\lambda_1$ and $\lambda_2$, set to 1 throughout the experiments.

% KL OBJECTIVE

The $cross$ objective diversifies the predictions by minimizing the negative mutual cross-entropy of any two models:
% \setlength{\abovedisplayskip}{5pt}
% {\small
\begin{equation*}
\mathcal{L}^{\text{cross}}_{\text{reg}} = - \frac{1}{N_m(N_m-1)} \sum_{i \neq j} \frac{\text{CE}(f_i, \text{argmax}(f_j)) + \text{CE}(f_j, \text{argmax}(f_i))}{2}
\end{equation*}
% }
% \setlength{\belowdisplayskip}{5pt}

Finally, the $kl$ objective aims at maximizing the $kl$ divergence between the output distributions of any two models, while the $div$ diversification objective is adapted from~\citep{lee2022diversify} and encourages diversity by minimizing the mutual information of any two models' predictions:

\begin{minipage}{\linewidth}
    \hspace{0pt}
    \begin{minipage}{0.495\linewidth}
        \begin{equation*}
            \mathcal{L}^{kl}_{\text{reg}} = - \frac{1}{N_m(N_m-1)} \sum_{i \neq j} D_{KL}(f_i || f_j)
        \end{equation*}
    \end{minipage}
    \hspace{-10pt}
    \begin{minipage}{0.495\linewidth}
        \begin{equation*}
            \mathcal{L}^{\text{div}}_{\text{reg}} = \frac{1}{N_m(N_m-1)} \sum_{i \neq j} D_{KL}(p(f_i, f_j) || p(f_i))
        \end{equation*}
    \end{minipage}
\end{minipage}
% Finally, the $div$ diversification objective is adapted from~\citep{lee2022diversify}, and can be summarized as:
% \setlength{\abovedisplayskip}{5pt}
% \begin{equation}
% \begin{aligned}
% \mathcal{L}^{\text{div}}_{\text{reg}} = \sum_{i \neq j} D_{KL}(p(f_i, f_j) || p(f_i) \otimes p(f_j)) + \sum_i D_{KL}(p(f_i) || p(y))
% \mathcal{L}^{\text{div}}_{\text{reg}} = \sum_{i \neq j} D_{KL}(p(f_i, f_j) || p(f_i)
% % \end{aligned}
% \end{equation}
% encouraging diversity by minimizing the mutual information of any two models' predictions. 

% -------------------------------------------------------------
% -------------------------------------------------------------
% -------------------------------------------------------------
% ---------------------   RESULTS    -----------------------
% -------------------------------------------------------------
% -------------------------------------------------------------
% -------------------------------------------------------------
% -------------------------------------------------------------

\begin{figure}
\centering
\vspace{-20pt}
\includegraphics[width=.9\linewidth]{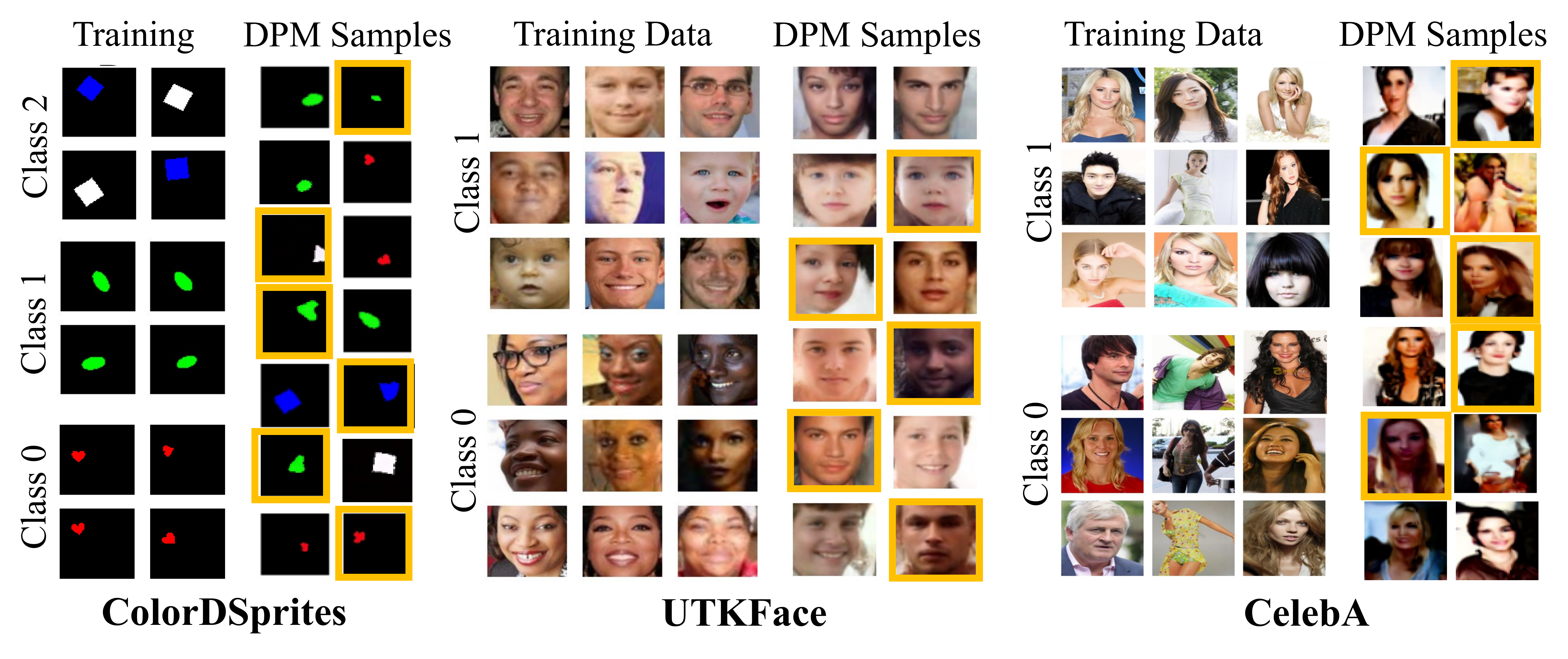}
\caption{DPM training and counterfactual generation. 
While training on images showcasing a correlated set of features 
(left columns), DPM samples at appropriate fidelity levels can generate novel objects beyond the observed feature combinations (marked right-hand side images).
} 
\label{fig:diff_sampling}
\vspace{-10pt}
\end{figure}

\section{Results} \label{sec:results}

\subsection{Implementation} \label{sec:exp_design}
To investigate the objectives of this study, we apply the WCST-ML framework~\citep{scimeca2022shortcut} on three datasets, namely ColorDSprites, UTKFace and CelebA. Each dataset is thus divided into a training set, comprising fully correlated feature-labels groups (\autoref{fig:diffdiv}), and as many test sets as there are label classes. For ColorDSprites, we consider the features of ${color, orientation, scale, shape\}}$, in UTKFace we consider the features ${\{ethnicity, gender, age\}}$, while in CelebA we consider the features ${\{light skin, oval face\}}$. We provide more implementation details in Suppl \autoref{sup:sec:methods}.

For the first stage we train three DPMs on the datasets with fully correlated features, each including respectively 34998, 1634 and 2914 feature-correlated samples. We generate $\approx 100$k samples from DPMs trained at varying number of epochs between 1 to 1.2K, to be used for ensemble diversification and analysis. We perform no post-processing or pruning of the generated samples, and instead wish to observe the innate ability of DPMs to generate samples beyond the training distribution.

In the second stage, we separately train ensembles for each combination of dataset and diversification objective, and for each, we perform ablation studies by applying the diversification objective to the data generated by the DPMs at different fidelity levels. For comparison, we also consider ensemble diversification with real \emph{ood} data, as well as a standard ensemble baseline with no diversification.

% In summary, the full analysis in this work involved the training of over $\approx 16,000$ ResNet-18 classifiers (each $\approx 11$million parameters), sectioned into 160 ensembles (one for each diffusion-objective pair and baselines), and 31 diffusion models (UNet $\approx 35$million parameters).

\subsection{$\text{Diffusion Counterfactual Sampling}$}\label{sec:res:diff_disentanglement}

% \subsubsection*{DPMs Exhibit Generalization Capabilities Under Correlation.}

\paragraph{DPMs Exhibit Generalization Capabilities Under Correlation:} We show in \autoref{fig:diff_sampling} the training samples for both datasets (left halves), as well as samples from the DPMs (right halves) trained for 25, 800, and 1000 epochs, respectively for each dataset. We observe how sampling from the trained DPMs generates previously unseen feature combinations, despite the correlated coupling of features during training (e.g. $\langle green/white, heart \rangle$, $\langle child, female \rangle$ or $\langle \neg light\ skxin, oval\ face \rangle$), suggesting a potential for DPMS to transcend surface-level statistics of the data, extending previous findings to the special case of fully correlated input features in the target distribution. 

\paragraph{Early Stopping to Capture Diffusion \emph{ood} Sampling Capabilities:} To understand under which conditions sampling from DPMs can lead to the generation of samples displaying unseen feature combinations, we examine the fraction of \emph{ood} samples generated by DPMs at different fidelities (\S \ref{sec:exp_design}).
\begin{wrapfigure}[18]{r}{0.49\textwidth}
\vspace{-18pt}
\centering
\includegraphics[width=.9\linewidth]{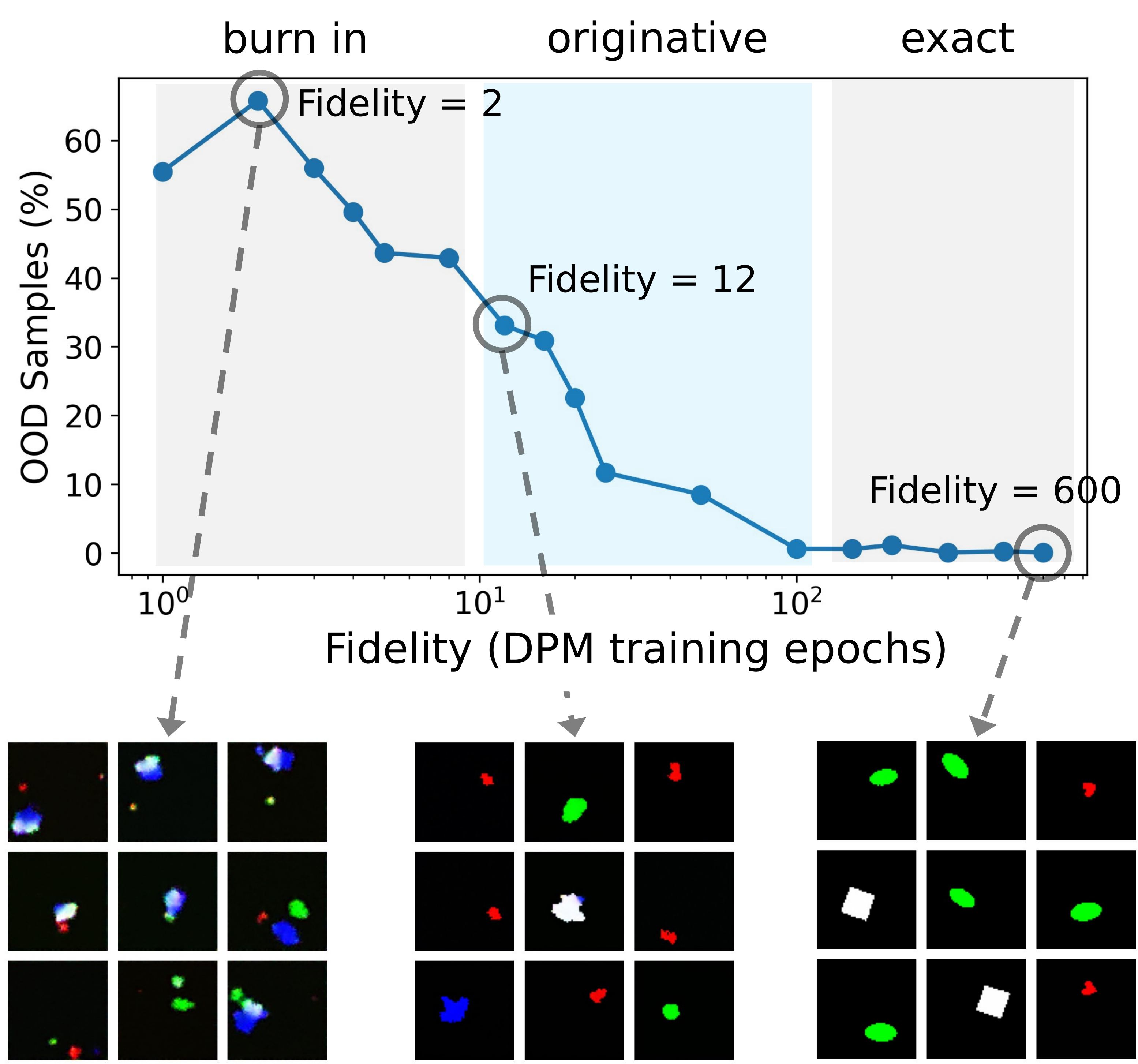}
\caption{\emph{ood} sample frequency for DPMs trained at different fidelities on ColorDSprites.}
\label{fig:fidelity_sampling}
\end{wrapfigure} 
\begin{wrapfigure}[12]{r}{0.35\textwidth}
\vspace{-48pt}
\includegraphics[width=.93\linewidth]{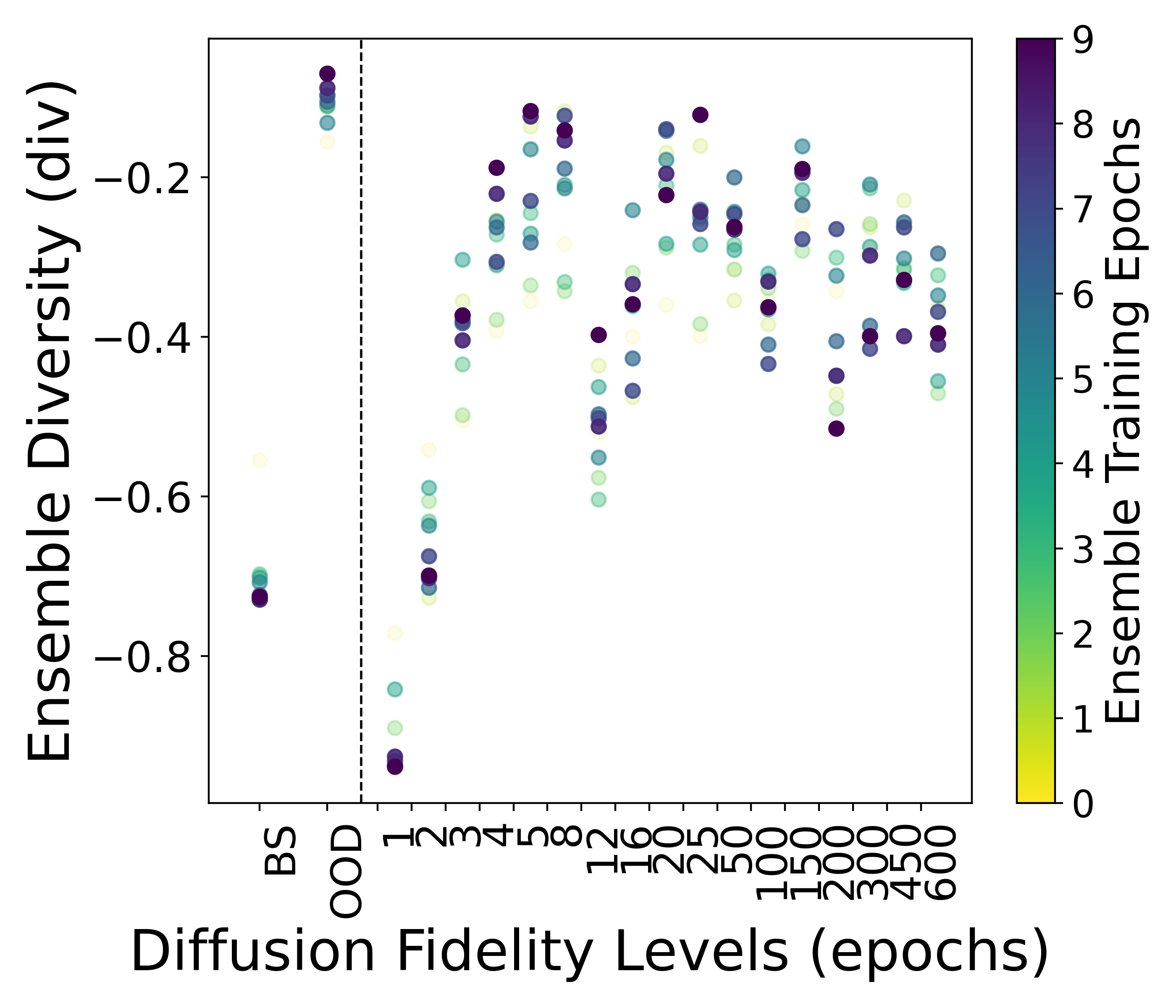}
\caption{Output diversity using samples from diffusion models at different fidelities, compared to using real \emph{ood} samples (ood), or without diversification (BS) in ColorDSPrites. All results in \autoref{sup:fig:fidelity_vs_diversity}. } 
\label{fig:fidelity_vs_diversity} 
\end{wrapfigure}

\vspace{-20pt}
We use the {ColorDSprites} for the disentangled
nature of its features, and the presence of all feature combinations in the data. 
To measure the fraction of \emph{ood} samples, we train a near-perfect oracle on the full dataset, and classify samples based on the \emph{diagonal} and \emph{off-diagonal} sets in WCST-ML \citep{scimeca2022shortcut}. \autoref{fig:fidelity_sampling} shows the fraction of \emph{ood} samples generated at varying levels of training fidelities. We identify at least three qualitative different intervals. An initial \emph{burn-in} interval, characterized by a high frequency of \emph{ood} generated samples, but which fails to capture the manifold of the data; an \emph{originative} interval, where we observe a reduced number of \emph{ood} samples, in favor for a generative distribution more aligned with the data to represent; and an \emph{exact} interval, where the DPM's ability to almost perfectly represent the data comes at the cost of novel emergent feature mixtures. In the context of ensemble diversification and bias mitigation, the \emph{originative} interval is posed to provide the necessary information to break the simplicity shortcuts, while providing effective disagreement signals.  Interestingly, we will later show that in \emph{DiffDiv}, low-fidelity samples can induce appropriate diversification. See Suppl. \S \ref{sup:sec:res} for further insights into DPM-early stopping with ensemble diversification criteria.

\subsection{Diffusion-guided Ensemble Diversity} \label{sec:diff_guided_ens_diversity}

We test whether diffusion counterfactuals can lead to ensemble diversification. We train ensembles comprising 100 ResNet-18 models on ColorDSprites, UTKFace, and CelebA. We train the ensembles with a cross-entropy loss on the correlated diagonal training data, as well as an additional diversification loss, computed on a separate diversification set. In our experiments, we will refer to `\emph{ood}' to the case where the data for disagreement is comprised of left-out, feature-uncorrelated, samples from the original dataset; to `diffusion' when the samples are generated by a DPM; and to `baseline' when no diversification objective is included. We tune $\gamma$ by grid-search to provide comparable experiments under different objectives (see Suppl. \S \ref{sup:sec:gamma})

% \subsubsection*{DPM Fidelity Significantly Impacts Diversification.}

\paragraph{DPM Fidelity Significantly Impacts Diversification:} \hfill\\We consider the case where the diversification objective is computed on samples generated by a DPM at different fidelity levels. We show in \autoref{fig:fidelity_vs_diversity} the diversity achieved with the \emph{div} diversification objective when using diffusion generated data at different fidelities. We observe a significant variation in the diversity of the ensemble predictions, from the low-diversity baseline (BS), trained without a diversification objective, to the high diversity achieved with \emph{ood} samples. From the figure, we observe how low fidelities provide little use for diversification, leading to homogeneous ensemble predictions. The DPM trained on ColorDSprites, however, quickly fits the synthetic data distribution, providing diversification samples on par with \emph{ood} data. Ultimately, excessive diffusion training leads to \emph{id} sample generation and lower diversification performance. We report analogous results for the UTKFace and CelebA dataset in Suppl \autoref{sup:fig:fidelity_vs_diversity}. Importantly, we find that appropriate DPM training and early stopping procedures are key to achieve diversity. We expand on early stopping signals for DPM training in Suppl. \ref{sup:sec:diversification}.

% \subsubsection*{Diffusion-guided Diversity Leads to Comparably Diverse Ensembles.}

%%%%%%%%%%%%%%%%%%%%%%%%%%%%%%%
%%%%%%%%%%%%%%%%%%%%%%%%%%%%%%%
%%%%%%%%%%%%%%%%%%%%%%%%%%%%%%%
%%%%%%%%%%%%%%%%%%%%%%%%%%%%%%%
%%%%%%%%%%%%%%%%%%%%%%%%%%%%%%%

\begin{table}
\vspace*{-2em}
\caption{Comparison between diversification with  \emph{ood}  samples and DiffDiv with model disagreement on five objectives. The feature columns report the fraction of models (not accuracy) attending to the respective features. The final column reports the average validation accuracy on the original training data. The shortcut feature for each dataset is highlighted in bold.}
\centering
\resizebox{1\linewidth}{!}{
\begin{tabular}{@{}lccccccccccccc}\toprule
Dataset $\rightarrow$ & \multicolumn{6}{c}{ColorDSprites} & \multicolumn{4}{c}{UTKFace} & \multicolumn{3}{c}{CelebA} \\\cmidrule(lr){2-7}\cmidrule(lr){8-11}\cmidrule(lr){12-14} 
Algo $\downarrow$ Feat. & Obj. $\downarrow$  $\rightarrow$  & \textbf{Color} ($\downarrow$)  & Orient.  & Scale  & Shape  & Acc. ($\uparrow$)   
        & Age  & \textbf{Ethnicity} ($\downarrow$)  & Gender  & Acc. ($\uparrow$)  
        & Oval Face & \textbf{Pale Skin} ($\downarrow$) & Acc. ($\uparrow$)\\\midrule

Baseline & -           & $1.00$          & $0.00$                & $0.00$          & $0.00$          & $1.000$\std{$0.00$}   
         & $0.00$          & $1.00$              & $0.00$             & $0.920$\std{$0.02$}  
         & $0.00$          & $1.00$            & $0.857$\std{$0.01$} \\\midrule

\multirow{5}{*}{OOD}    
             & Cross & $0.99$          & $0.00$                & $0.01$          & $0.00$          & $0.865$\std{$0.17$}   
             & $0.01$          & $0.71$              & $0.28$             & $0.865$\std{$0.04$}  
             & $0.01$          & $0.99$            & $0.751$\std{$0.07$}  \\
             & Div   & $0.86$          & $0.01$                & $0.11$          & $0.02$          & $0.818$\std{$0.22$}   
             & $0.00$          & $0.94$              & $0.06$             & $0.859$\std{$0.03$}  
             & $0.00$          & $1.00$            & $0.843$\std{$0.03$}  \\
             & KL    & $0.91$          & $0.02$                & $0.07$          & $0.00$          & $0.822$\std{$0.21$}   
             & $0.05$          & $0.76$              & $0.19$             & $0.818$\std{$0.06$}  
             & $0.00$          & $1.00$            & $0.812$\std{$0.04$}  \\
             & L1    & $0.91$          & $0.00$                & $0.08$          & $0.01$          & $0.813$\std{$0.20$}   
             & $0.02$          & $0.62$              & $0.36$             & $0.847$\std{$0.07$}  
             & $0.14$          & $0.86$            & $0.724$\std{$0.11$}  \\
             & L2    & $0.84$          & $0.02$                & $0.13$          & $0.01$          & $0.729$\std{$0.23$}   
             & $0.03$          & $0.63$              & $0.34$             & $0.798$\std{$0.11$}  
             & $0.12$          & $0.88$            & $0.651$\std{$0.10$}  \\\midrule

\multirow{5}{*}{\textbf{DiffDiv (ours)}}      
             & Cross & $0.96$          & $0.00$                & $0.04$          & $0.00$          & $0.856$\std{$0.16$}   
             & $0.00$          & $0.94$              & $0.06$             & $0.836$\std{$0.05$}  
             & $0.01$          & $0.99$            & $0.745$\std{$0.10$}  \\
             & Div   & $0.94$          & $0.00$                & $0.06$          & $0.00$          & $0.916$\std{$0.13$}   
             & $0.00$          & $0.98$              & $0.02$             & $0.826$\std{$0.05$}  
             & $0.00$          & $1.00$            & $0.857$\std{$0.01$}  \\
             & KL    & $0.89$          & $0.01$                & $0.10$          & $0.00$          & $0.786$\std{$0.20$}   
             & $0.00$          & $0.94$              & $0.06$             & $0.837$\std{$0.06$}  
             & $0.04$          & $0.96$            & $0.672$\std{$0.07$}  \\
             & L1    & $0.89$          & $0.00$                & $0.09$          & $0.02$          & $0.784$\std{$0.20$}   
             & $0.00$          & $0.77$              & $0.23$             & $0.816$\std{$0.11$}  
             & $0.08$          & $0.92$            & $0.659$\std{$0.12$}  \\
             & L2    & $0.93$          & $0.02$                & $0.03$          & $0.02$          & $0.762$\std{$0.22$}   
             & $0.01$          & $0.83$              & $0.16$             & $0.757$\std{$0.12$}  
             & $0.09$          & $0.91$            & $0.650$\std{$0.11$}  \\\bottomrule

\end{tabular}
}
\label{tab:diversity:all}
\vspace{-10pt}
\end{table}

%%%%%%%%%%%%%%%%%%%%%%%%%%%$$
%%%%%%%%%%%%%%%%%%%%%%%%%%%%%%%
%%%%%%%%%%%%%%%%%%%%%%%%%%%%%%%
%%%%%%%%%%%%%%%%%%%%%%%%%%%%%%%
%%%%%%%%%%%%%%%%%%%%%%%%%%%%%%%

\begin{figure}
\centering
\includegraphics[width=\linewidth]{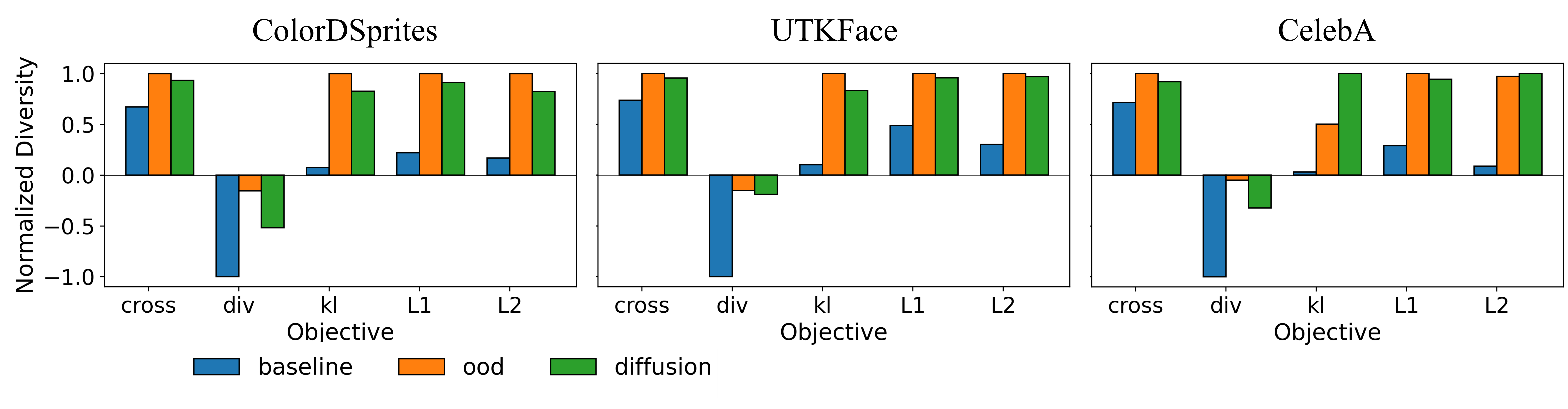}
\caption{Average normalized ensemble diversity comparisons across metrics (higher is better).} 
\label{fig:diff_ood_comparison}
% \vspace{-10pt}
\vspace{-10pt}
\end{figure}

\paragraph{Diffusion-guided Diversity Leads to Comparably Diverse Ensembles:} Given the difficulty and cost of collecting auxiliary \emph{ood} data, we wish to compare the level of diversification achieved by using DPM counterfactuals, as opposed to real \emph{ood} data. Based the results in \autoref{fig:fidelity_vs_diversity}, we choose samples drawn from DPMs trained for 100, 800 and 1200 epochs for ColorDSprites, UTKFace and CelebA respectively, expected to be especially useful for diversification.
In \autoref{fig:diff_ood_comparison} we report the objective-wise normalized diversity achieved in each scenario by the ensemble. We find that diffusion counterfactuals lead to comparable diversification performance to real \emph{ood} samples. In the figure, the diffusion-led diversity is almost always within 5\% from the metrics achieved when using pure \emph{ood} samples, both typically 
\begin{wraptable}[10]{r}{0.38\textwidth}
\vspace{-5pt}
% \begin{table}
% \begin{minipage}{.38\textwidth}
\centering
\caption{Average change in ensemble accuracy over the averted cues following diversification (reported in \%).}
\resizebox{\linewidth}{!}{
\begin{tabular}{@{}lccc}\toprule
Obj. $\downarrow$ Dataset $\rightarrow$ & Color & Face & CelebA \\\midrule

Cross     & 5.17 & 3.37 & 2.00 \\
Div       & 8.16 & 3.61 & 0.45 \\
KL        & 2.87 & 4.54 & 2.02 \\
L1       & 3.31 & 3.61 & 1.30 \\
L2       & 4.67 & 5.04 & 1.34 \\\bottomrule
\end{tabular}
}
\label{tab:increase_in_accuracy}
% \end{minipage}%
% \end{table}
\end{wraptable}
over 50\% higher than the baseline. 

\paragraph{Ensemble Diversification Breaks Simplicity Biases:} By WCST-ML, we can test each model's bias to a cue by testing the model's output on purposefully designed test sets~\citep{scimeca2022shortcut}. We report in \autoref{tab:diversity:all} the fraction of models attending to specific cues, when trained on real \emph{ood} data and diffusion-generated samples (\emph{DiffDiv}). A model was deemed to be attending to a cue if its validation accuracy on the cue-specific \emph{ood} WCST-ML dataset was highest relative to all other features. Firstly, training with no diversification on fully correlated data yields especially biased ensembles, solely attending to the cues of $colore$, $ethnicity$, and $pale\ skin$. Notably, upon introducing the diversification objectives, we observed a perceptible shift in the models' behavior, averting their attention from primary, easy-to-learn, cues, to other latent cues present within the data. 
% Among the objectives considered, $kl$, $L_1$, and $L_2$ exhibited the highest cue diversity, catalyzing the ensemble to distribute attention across multiple cues. However, this is at the expense of a marked drop in the average ensemble performance. Conversely, the \emph{div} and \emph{cross} objectives yielded milder diversification, focusing on the next readily discernible cues: \emph{scale} in ColorDSprites, \emph{gender} in UTKFace, and \emph{oval face} in CelebA; while maintaining a generally higher ensemble validation performance. 
Similarly to \autoref{fig:diff_ood_comparison}, we find the  diversification induced by DiffDiv in \autoref{tab:diversity:all} is largely comparable with \emph{ood} data on ColorDSprites, with respectively up to $11\%$ and $14\%$ of the models averting their attention from the main `color' cue in ColorDSprites; up to $23\%$ and $38\%$ of the models averting their attention to the ethnicity cue in UTKFace; and up to $9\%$ and $14\%$ of the models averting their attention to the ethnicity cue in CelebA. Furthermore, we report the change in ensemble accuracy over the averted cues following diversification in \autoref{tab:increase_in_accuracy}, confirming a positive and significant, increase for all experiments.

% -------------------------------------------------------------
% -------------------------------------------------------------
% -------------------------------------------------------------
% -----------------   DISCUSSION/CONCLUSION    ----------------
% -------------------------------------------------------------
% -------------------------------------------------------------
% -------------------------------------------------------------
% -------------------------------------------------------------

\section{Discussion and Conclusion}

Shortcut learning is a phenomenon undermining the performance and utility of deep learning models, where easy-to-learn cues are preferentially learned for prediction. We propose \emph{DiffDiv}, a framework to achieve shortcut learning mitigation via ensemble diversification on low-fidelity DPM counterfactuals. We show that DPMs can generate novel feature combinations even when trained on images displaying correlated cues. We identify a DPM training stage (\emph{originative}) that is especially useful for ensemble diversification, and show that diffusion-guided diversification can lead to comparable ensemble diversity without the need for expensive \emph{ood} data collection.


\subsubsection*{Acknowledgments}
Luca Scimeca acknowledges funding from Recursion Pharmaceuticals. Damien Teney is partially supported by an Amazon Research Award. The research was enabled in part by computational resources provided by the Digital Research Alliance of Canada (\url{https://alliancecan.ca}), Mila (\url{https://mila.quebec}), and NVIDIA.

% \newpage
\bibliographystyle{iclr2025_conference}
\bibliography{main}

\newpage

\clearpage
\setcounter{page}{1}

% -- additional by luca --
\setcounter{section}{0}
\renewcommand{\thesection}{S\arabic{section}}
\setcounter{table}{0}
\renewcommand{\thetable}{S\arabic{table}}%
\setcounter{figure}{0}
\renewcommand{\thefigure}{S\arabic{figure}}%
%%%%%%%%%%%%%%%%%%%%%%%%%
% \section{Preliminaries}

\section{Related Work} \label{sup:sec:related_work}
\subsection{Overcoming the Simplicity Bias}

\paragraph{Deliberate de-bias:} To overcome the simplicity bias, copious literature has explored methodologies to avoid or mitigate shortcut cue learning when labels for the shortcut cues were present in the training data~\citep{Repair2019, kirichenko2023last, EstimatingAndMitigating2019, LearningNotToLearn2019, Sagawa2020Distributionally, lee2021learning}. The access to shortcut signals is, however, a critical limitation, as these are generally hard or impossible to obtain. 

\paragraph{Data augmentation:} Data augmentation methodologies have proven useful in the generation of \emph{bias-conflicting} or bias-free samples~\citep{kim2021biaswap, lim2023biasadv, jung2023fighting}, or to augment underrepresented subgroups therein~\citep{wang2020deep, mondal2023minority}, leading to less biased predictions. Although an important research direction, assumptions on the nature of the biases are here still necessary, and the rejection of selected cues for prediction may not always lead to improved generalization, notably when the downstream task is aligned with said cues. 

\paragraph{Diversification:} A different approach to this problem has been to enforce the use of a diverse set of signals for prediction. The use of ensembles has been one such method, where models' diversity would lead to bias mitigation and improved generalization. Different weight initialization and architectures have previously been shown to be ineffective in the presence of strong shortcut biases~\citep{scimeca2022shortcut}. Methods ensuring mutual orthogonality of the input gradients have proven more effective, driving models to attend to different locations of the input space for prediction~\citep{ross2020ensembles, teney2022evading, teney2022predicting, nicolicioiu2023learning}. These input-centric methods, however, may be at a disadvantage in cases where different features must attend to the same area of the input space. Instead, a different approach has been via the diversification of direct ensemble model predictions. This approach hinges on the availability of --unlabelled-- out-of-distribution (\emph{ood}) auxiliary samples that are, at least in part, free of the same shortcuts as the original training data. Through a diversification objective, the models are then made to disagree on these \emph{ood} samples, while maintaining performance on the original data, effectively fitting functions with different extrapolation behaviors~\citep{lee2022diversify, pagliardini2022agree, scimeca2023leveraging, lin2023spurious}. 
Even in this case, the auxiliary \emph{ood} data dependency poses limitations, as this is often not readily accessible, and can be costly to procure.

\subsection{DPM Modelling and Generalization}

In recent years, Diffusion Probabilistic Models have emerged as a transformative generative tool, spearheading progress in the generative domain across various applications, including vision \citep{ho2020denoising, song2019generative}, efficient sampling methods \citep{sendera2024diffusion}, and text \citep{gong2022diffuseq, venkatraman2024amortizing} and even control tasks \citep{venkatraman2024amortizing}. Numerous studies have underscored their prowess in generating synthetic images, which can then be harnessed to enrich datasets and bolster classification performance~\citep{sariyildiz2023fake, azizi2023synthetic, yuan2022not, dunlap2023diversify, howard2023probing}.  

In several cases, DPMs have been shown to transcend the surface-level statistics of the data, making them invaluable in understanding data distributions and features~\citep{chen2023beyond, yuan2022not, wu2023uncovering}. Moreover, recent studies have indicated the ability of DPMs to achieve feature disentanglement via denoising reconstructions~\citep{kwon2022diffusion, wu2023uncovering, okawa2023compositional}. In particular, work in~\citep{wang2023concept} has shown how text-guided generative models can represent disentangled concepts, and how, through algebraic manipulation of their latent representations, it is possible to compositionally generalize to novel and unlikely combination of image features. 

Additional work has also shown strong inductive Diffusion biases during learning, which may explain some of these phenomena \citep{kadkhodaie2023generalization}. In our work, we test the edge case of DPMs trained with data exhibiting correlated input features. We find that when suitably trained, DPMs can still generate samples with novel feature combinations and that these can be leveraged for ensemble diversity.

\section{Supplementary Methods} \label{sup:sec:methods}
\subsection{Datasets}
In this work, we leverage three representative datasets: a color-augmented version of DSprites~\citep{dsprites17}, UTKFace~\citep{zhang2017age}, and CelebA~\citep{liu2015faceattributes}. 

\vspace{8pt} 
\noindent
\textbf{DSprites:} DSprites includes a comprehensive set of symbolic objects generated with variations in five latent variables: shape, scale, orientation, and X-Y position. We augment this dataset with a color dimension and remove redundant samples (e.g. due to rotations), resulting in \(2,862,824\) distinct images, which we refer to as ColorDSprites. ColorDSprites permits the examination of shortcut biases in a highly controlled setup. 

\vspace{8pt} 
\noindent
\textbf{UTKFace:} UTKFace provides a dataset of \(23,708\) facial images annotated with attributes like age, gender, and ethnicity. Unlike DSprites, UTKFace presents a real-world, less controlled setup to study bias. Its inherent complexity and diversity make it an ideal candidate for understanding the model's cue preferences when societal and ethical concerns are at stake. 

\vspace{8pt} 
\noindent
\textbf{CelebA:} CelebA is a large-scale dataset comprising \(202,599\) celebrity facial images annotated with \(40\) different attributes, including gender, age, and facial features. It provides a more extensive and diverse real-world dataset compared to UTKFace, offering rich variations in pose, background, and lighting. CelebA is commonly used for studying bias and fairness in models due to its attribute diversity and challenging conditions.

\begin{figure}
\centering
% \vspace{-10pt}
\includegraphics[width=.9\linewidth]{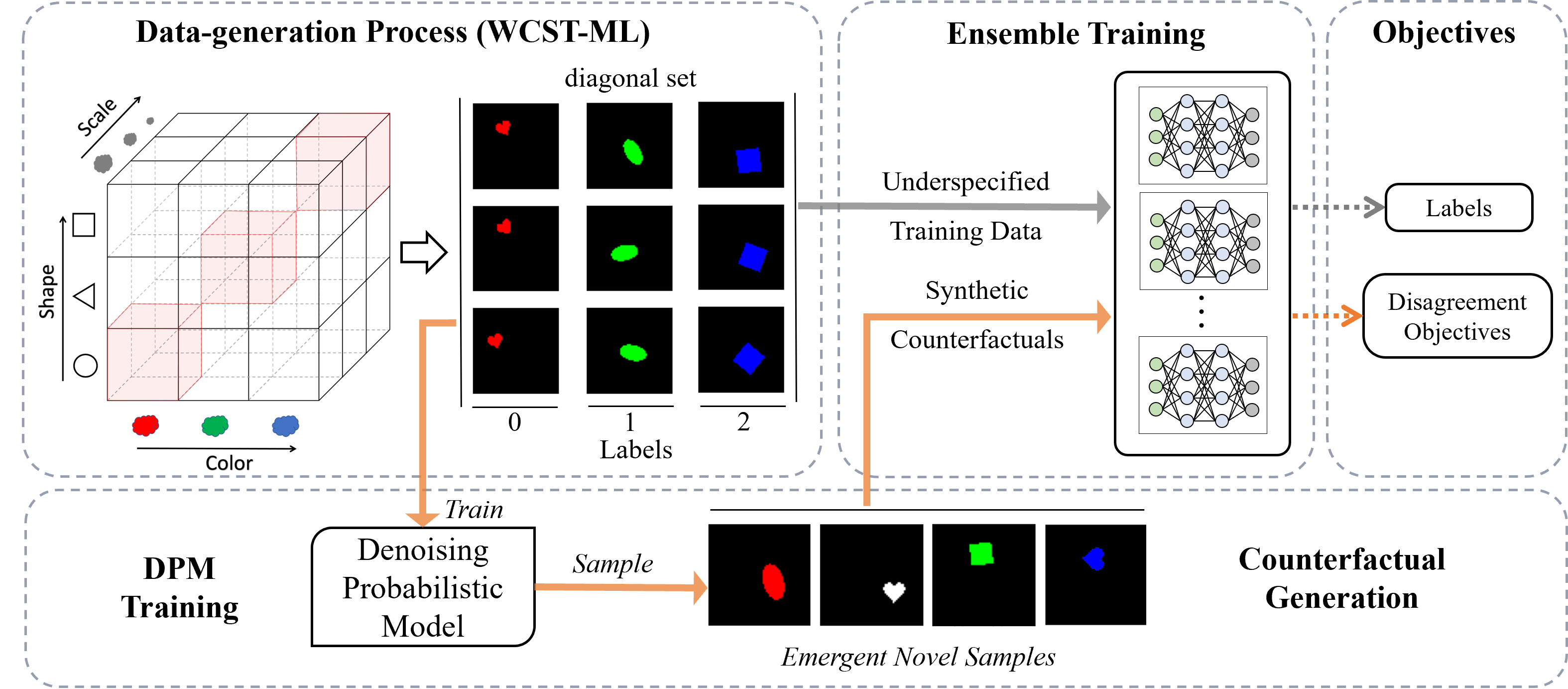}
\caption{We consider the WCST-ML setting~\citep{scimeca2022shortcut} as the experimental ground in our experiments. Each dataset is partitioned into training data where the task labels are perfectly correlated with the image input features (e.g. $\{$color, shape, and scale$\}$) and Test data, where samples are used to test a model's tendency to a feature over another. We use a random subset of Test Data for  \emph{ood}  experiments. } 
\label{sup:fig:framework}
% \vspace{-5pt}
\end{figure}

\subsection{Wisconsin Card Sorting Test for Machine Learners (WCST-ML)} \label{sec:wcst-ml}

% WCST-ML, what it is, how it decides train test splits, how it can provide a good test-bed for shortcut-cue learning
To isolate and investigate shortcut biases, we employ the Wisconsin Card Sorting Test for Machine Learners (WCST-ML), a method devised to dissect the shortcut learning behaviors of deep neural networks~\citep{scimeca2022shortcut}. We use the splits from WCST-ML to both train and evaluate the ensembles in this work, as it provides a systematic approach to creating datasets with multiple cues, designed to correlate with the target labels. Specifically, given $K$ cues $i_1, i_2, \ldots, i_K$, the method produces a \emph{diagonal} dataset $\mathcal{D}_\text{diag}$ where each cue $i_k$ is equally useful for predicting the labels $Y$, with the total number of classes $L = |Y|$. This level playing field is instrumental in removing the influence of feature dominance and spurious correlations, thereby allowing us to observe a model's preference for certain cues under controlled conditions. To rigorously test these preferences, WCST-ML employs the notion of \emph{off-diagonal} samples. These are samples where the cues are not in a one-to-one correspondence with the labels, but instead align with only one of the features under inspection. By evaluating a model's performance on off-diagonal samples, according to each feature, we can test and achieve an estimate of a model's reliance on the same. 
% Essentially, high unbiased accuracy for a cue $k$, denoted as $\text{acc}_k(f)$, signals that the model $f$ is heavily relying on that cue for its predictions.

\subsection{Operationalizing WCST-ML Across Datasets}

We follow the set-up in~\citep{scimeca2022shortcut} and construct a balanced dataset $\mathcal{D}_\text{diag}$, which includes a balanced distribution of cues, coupled with their corresponding off-diagonal test sets (one for each feature) \autoref{sup:fig:framework}. For both datasets, we define a balanced number of classes \(L\) for each feature under investigation. Where the number of feature values exceeds \(L\), we dynamically choose ranges to maintain sample balance with respect to each new feature class. For instance, for the continuous feature `age' in UTKFace, we dynamically select age intervals to ensure the same \(L\) number of categories as other classes, as well as sample balance within each category. We consider sets of features previously found to lead to strong simplicity biases. For ColorDSprites, we consider $K_{DS}=4$, features $\{color, orientation, scale, shape\}$, and $L=3$ as constrained by the number of shapes in the dataset. Within UTKFace we consider $K_{UTK}=3$, features $\{ethnicity, gender, age\}$, and $L=2$ as constrained by the binary classification on \emph{gender}. For CelebA we consider $K_{CL}=2$, features $\{light skin, oval face\}$, and $L=2$ as enforced by the binary labels on all features. For each dataset we create one \emph{diagonal} subset of fully correlated features and labels, available at training time, and $K_{DS}$, $K_{UTK}$ and $K_{CL}$ feature-specific \emph{off-diagonal} datasets to serve for testing the models' shortcut bias tendencies.

\subsection{DPM Training and Synthetic Counterfactual Generation}

We utilize Diffusion Probabilistic Models (DPMs) to generate synthetic data for our experiments. DPMs operate by iteratively adding or removing noise from an initial data point \( x \) through a stochastic process governed by a predefined noise schedule. We base our training regime on \citep{ho2020denoising}. As denoiser, we train a classic U-Net architecture with 4 down-sampling blocks and 4 up-sampling blocks with $\approx 9mil$ parameters for all datasets. We train the model with the objective in \autoref{eq:diff_objective} by iterating through the relevant dataset over a maximum of 1200 epochs. We use a vanilla Adam optimizer in all experiments.
All DPM schemes use a time discretization of 1000 steps.
To facilitate efficient sampling, we employ Denoising Diffusion Implicit Models (DDIM)~\citep{song2020denoising}, a first-order ODE solver for DPMs~\citep{salimans2022progressive, lu2022dpm}, utilizing a predictor-corrector scheme to minimize the number of sampling steps, lowering the final number to 250 during sampling in framework.

For each of the experiments, we use the trained DPM to generate a fixed dataset of 3000 synthetic counterfactuals, which is independently shuffled and batched-sampled for the diversification objective during ensemble training. We perform ablation studies considering a larger batch of synthetic counterfactuals (equal to the number of data points in each dataset) in \S \ref{sup:sec:res_diversification_all_ood}, with only marginal performance gains compared to the smaller set.

\section{Supplementary Results} \label{sup:sec:res}
\subsection{$\gamma$ Selection} \label{sup:sec:gamma}
\begin{figure}[t]
% \begin{wrapfigure}{r}{0.50\textwidth}
    
    \centering
    \begin{subfigure}[b]{.9\linewidth}
    \centering
        \begin{subfigure}[b]{.49\linewidth}
            \centering
            \includegraphics[width=.95\linewidth]{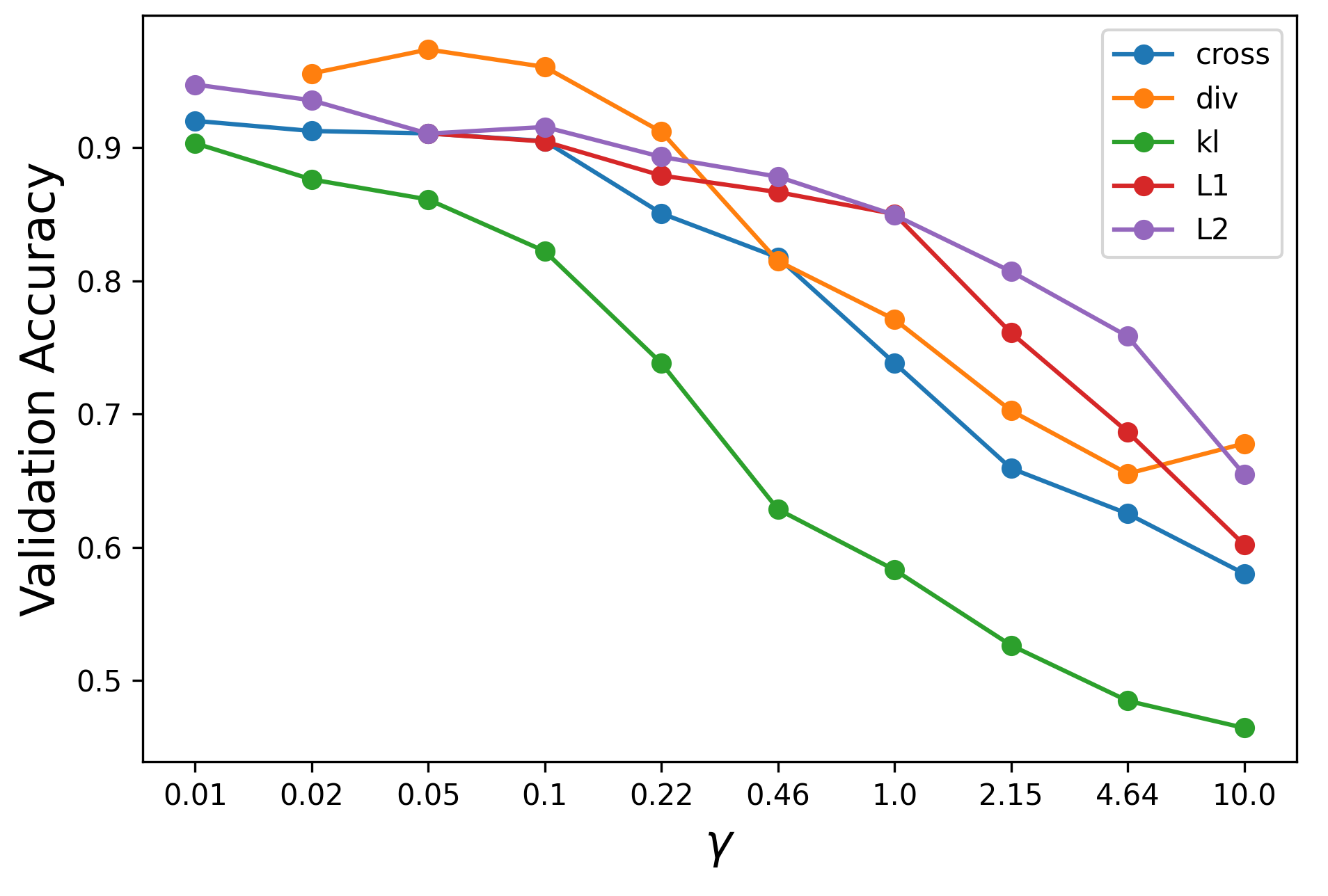}
        \end{subfigure}%
        \begin{subfigure}[b]{.49\linewidth}
            \centering
            \includegraphics[width=.95\linewidth]{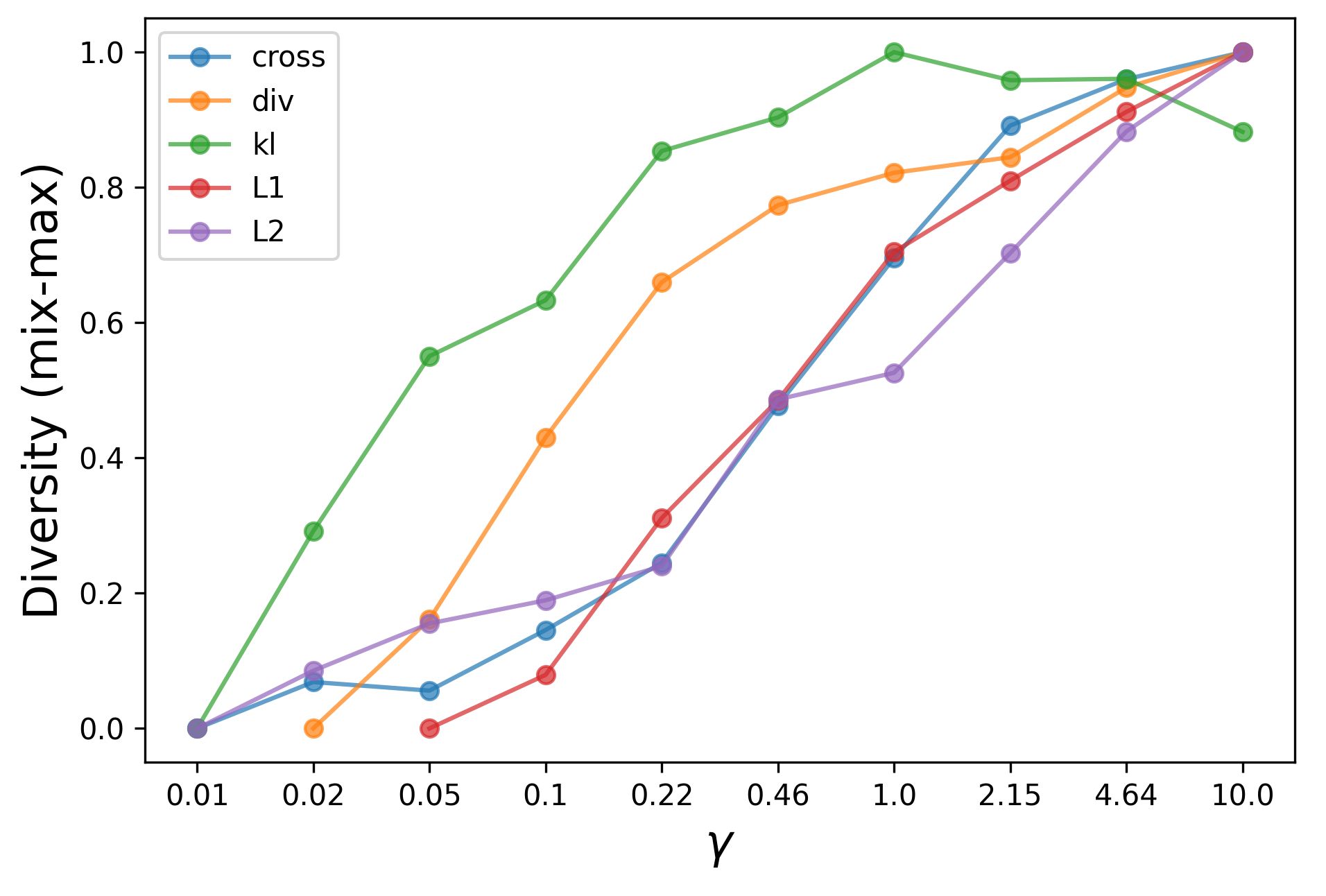}
        \end{subfigure}
    \caption{ColorDSprites}
    \label{fig:calibration_dsprites}
    \end{subfigure}
    \begin{subfigure}[b]{.9\linewidth}
    \centering
        \begin{subfigure}[b]{.49\linewidth}
            \centering
            \includegraphics[width=.95\linewidth]{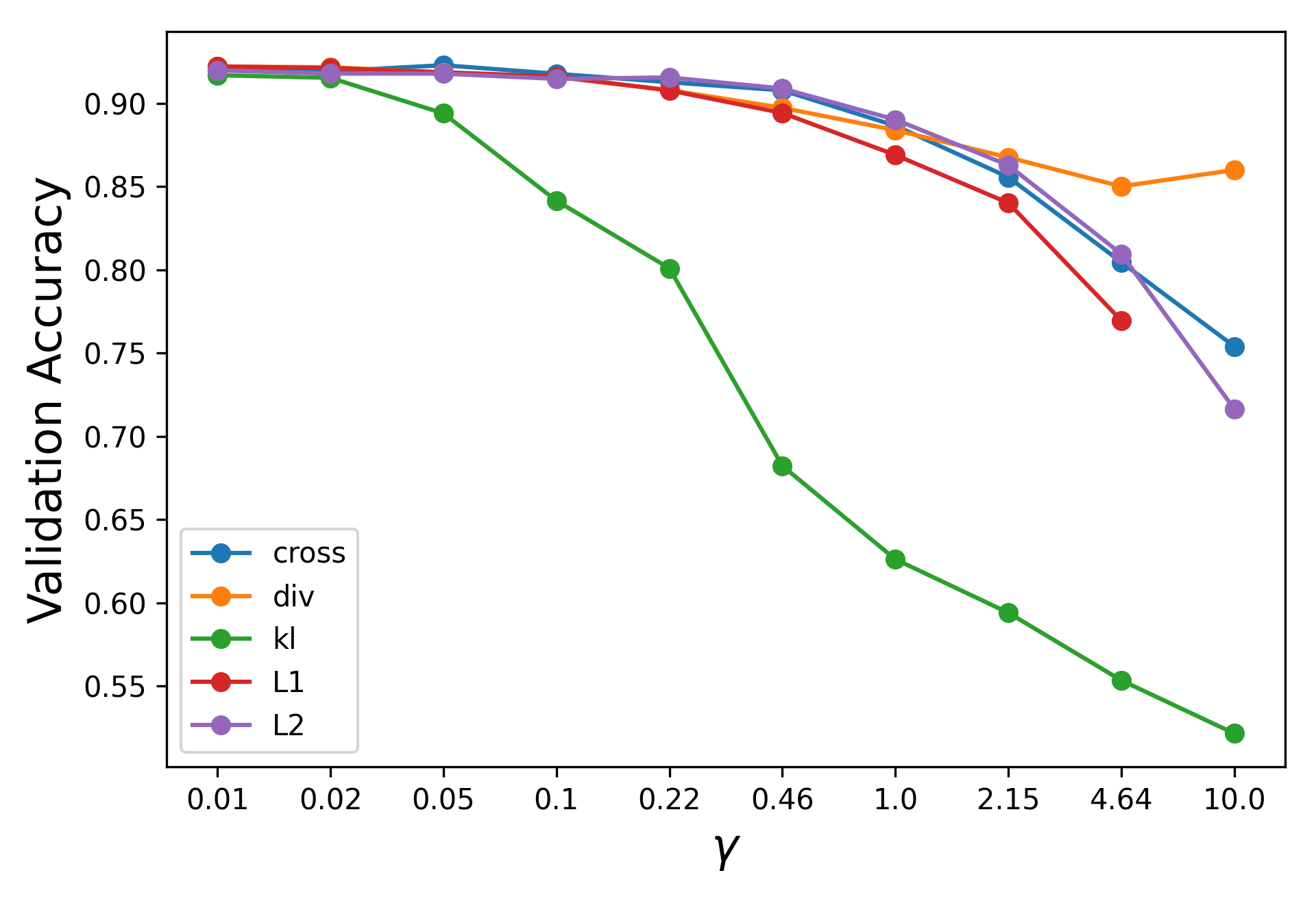}
        \end{subfigure}%
        \begin{subfigure}[b]{.49\linewidth}
            \centering
            \includegraphics[width=.95\linewidth]{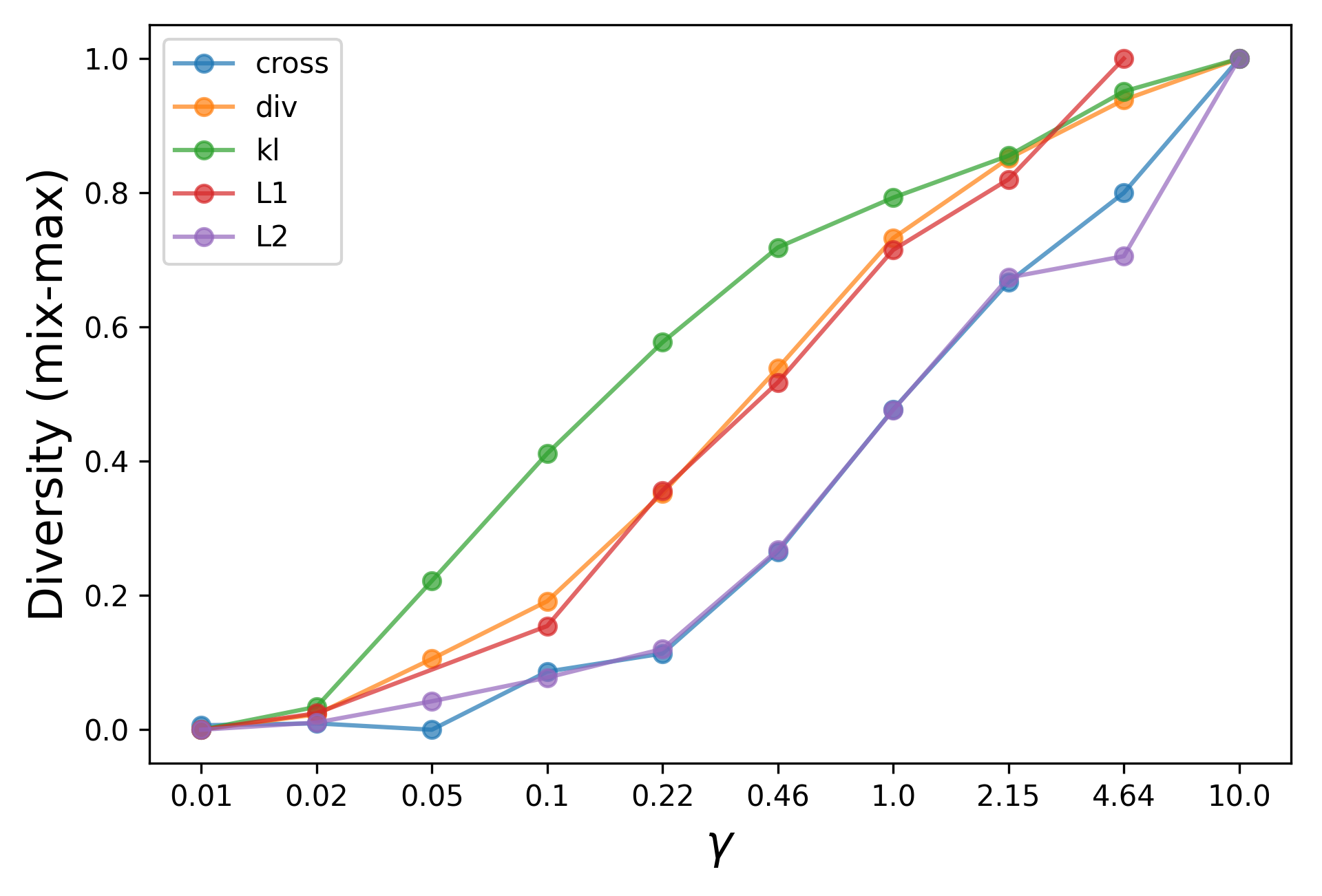}
        \end{subfigure}
    \caption{UTKFace}
    \label{fig:calibration_utkface}
    \end{subfigure}
    \begin{subfigure}[b]{.9\linewidth}
    \centering
        \begin{subfigure}[b]{.49\linewidth}
            \centering
            \includegraphics[width=.95\linewidth]{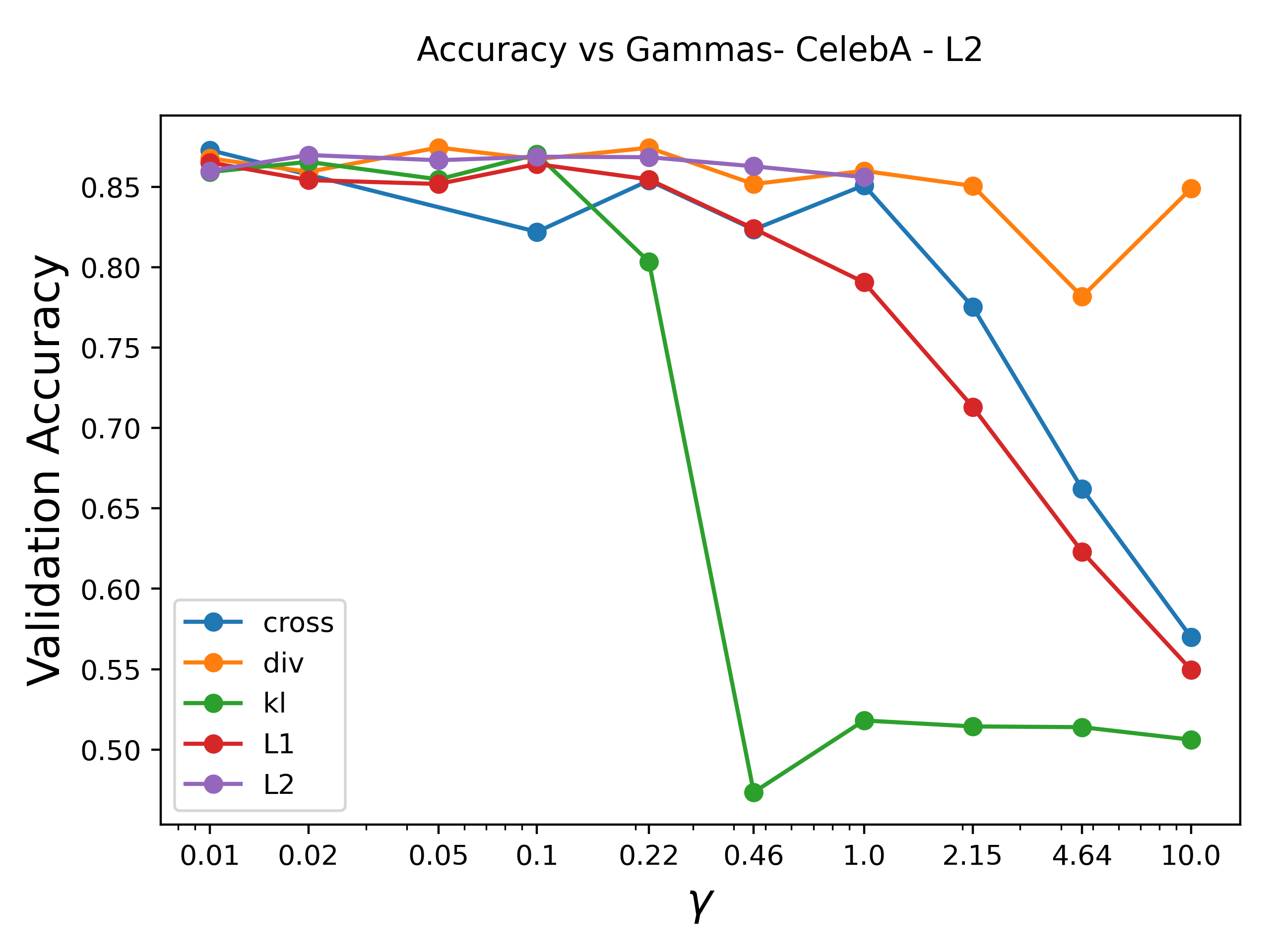}
        \end{subfigure}%
        \begin{subfigure}[b]{.49\linewidth}
            \centering
            \includegraphics[width=.95\linewidth]{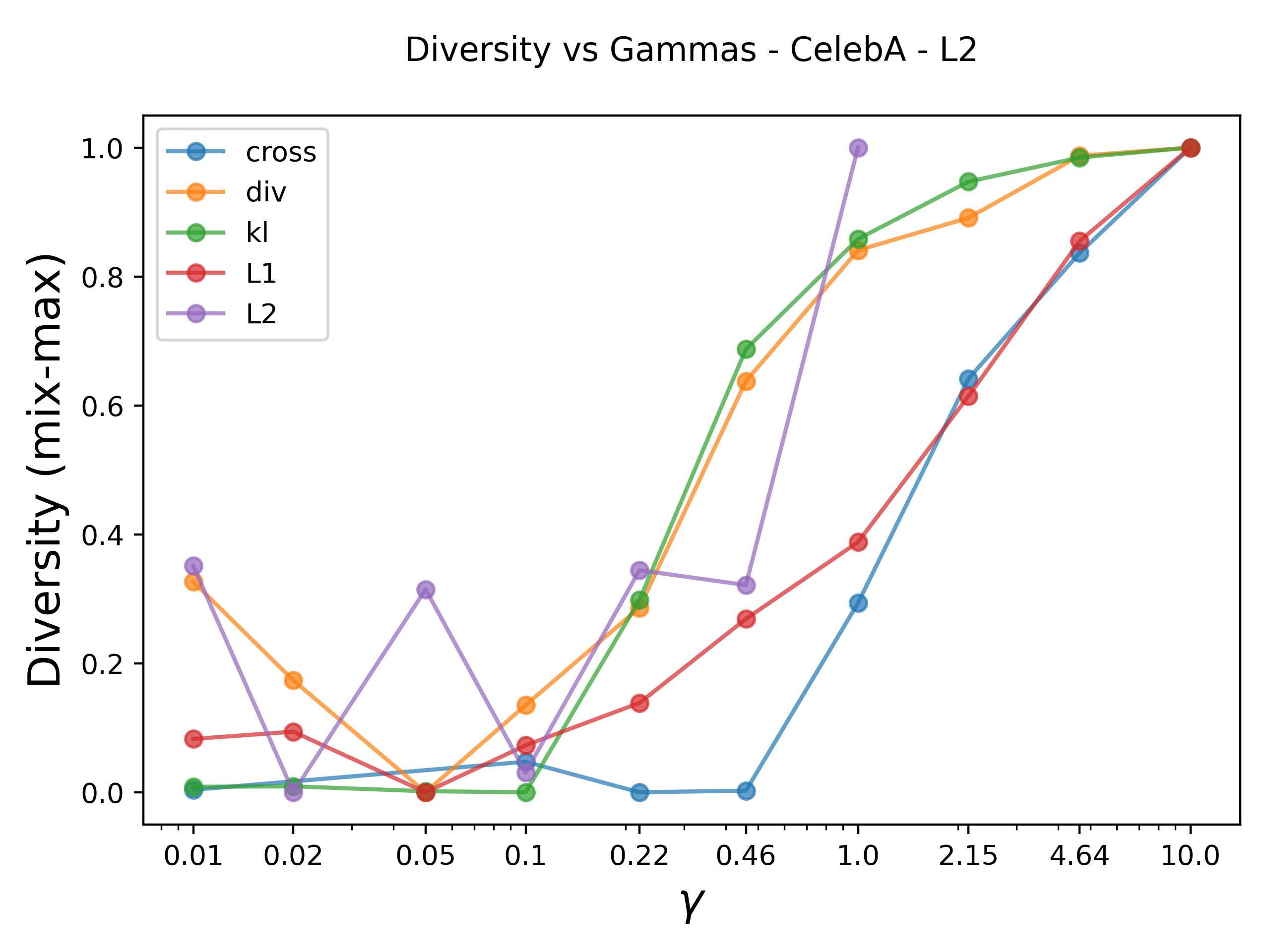}
        \end{subfigure}
    \caption{CelebA}
    \label{fig:calibration_utkface}
    \end{subfigure}
    \caption{Hyper-parameter search on the disagreement intensity ($\gamma$) for each diversification objective. For the main experiments, we select the values of $\gamma$ at the intersection of the accuracy-diversity trends by each objective (\autoref{suppl:tab:dis_hyper}).}
    \label{fig:calibration}
\end{figure}

We perform a hyper-parameter search to find the values of $\gamma$ for each diversification objective. We perform the search via the same methodology used in the \emph{ood} main experiments, i.e. by training an ensemble of 100 ResNet-18 models on the fully correlated \emph{diagonal} datasets, while diversifying a small subset of 30\% of the original training data, randomly sampled from the de-correlated left-out set. We consider $\gamma$ values ranging from $1e-3$ to $1e1$, and monitor both the validation ensemble accuracy as well as the predictive diversity on a separate de-correlated set of validation data. Figure \ref{fig:calibration} shows the performance of each metric for different values of $\gamma$. We select the values reported in \autoref{suppl:tab:dis_hyper} to be at the intersection of the accuracy and diversification trends for each model.

% \end{wrapfigure}

\subsection{Model Selection to Boost Diverse Ensemble Performance}

The mitigation of Shorcut leaning through diversification is generally known in the literature to suffer from a decrease in ensemble ID performance, as we also observe and study in our experiments. A prominent methodology to mitigate this phenomenon in the literature is ensemble model selection, where a subset of models is selected for final ensemble inference. We perform additional experiments to assess the degree by which model selection can aid in ensemble performance within the DiffDiv framework.
To ensure diversity in the final selection, we include in the selected subset any model showing shortcut-cue aversion from Table 1. Furthermore, we select additional models to reach a dynamic range between 15\% and 99\% of the original ensemble. 
We compare the performance of the ensemble before and after model selection in \autoref{tab:model_selection}.

\subsection{On the Influence of DPM Fidelity to Diversification} \label{sup:sec:diversification}
\begin{figure*}
\centering

\begin{subfigure}[b]{\textwidth}
    \centering
    
    \begin{subfigure}[b]{.195\columnwidth}
        \includegraphics[width=\linewidth]{figures/fidelity_diversity_trends/color_dis_div_fidelity_comparison.png}
        \caption{ColorDSprites} 
    \end{subfigure}
    \begin{subfigure}[b]{.195\textwidth}
        \centering
        \includegraphics[width=\linewidth]{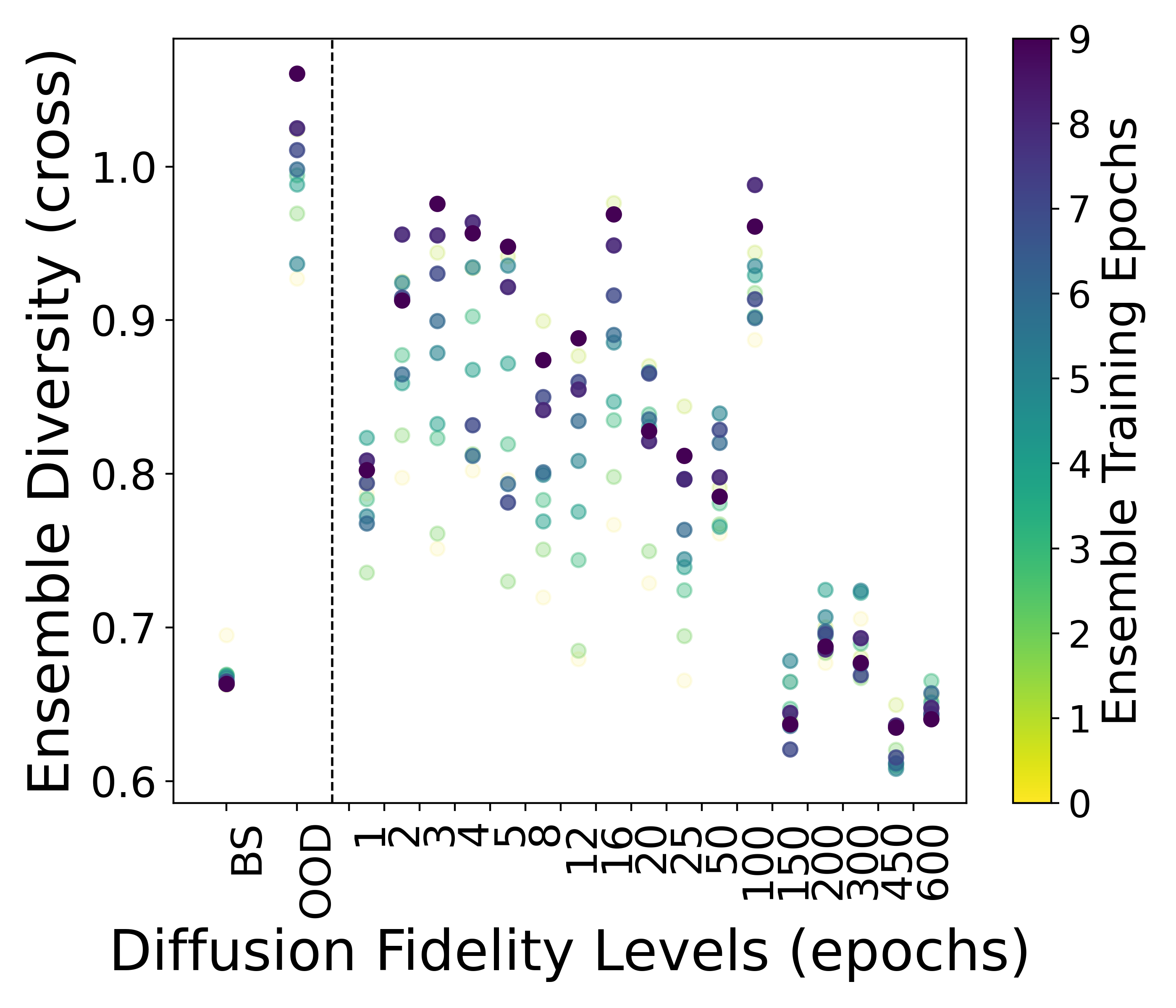}
        \caption*{cross} 
    \end{subfigure}
    \begin{subfigure}[b]{.195\textwidth}
        \centering
        \includegraphics[width=\linewidth]{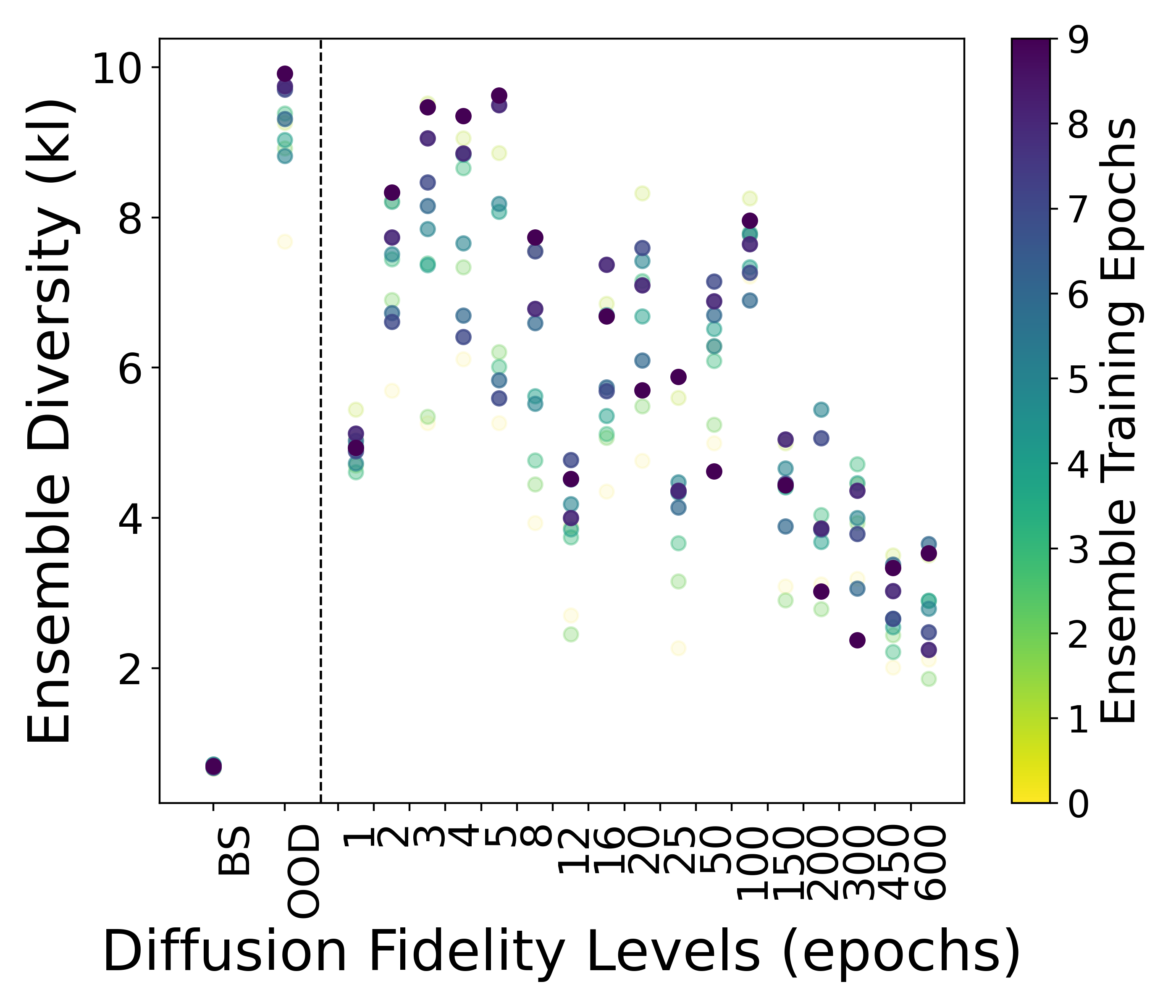}
        \caption*{kl} 
    \end{subfigure}
    \begin{subfigure}[b]{.195\textwidth}
        \centering
        \includegraphics[width=\linewidth]{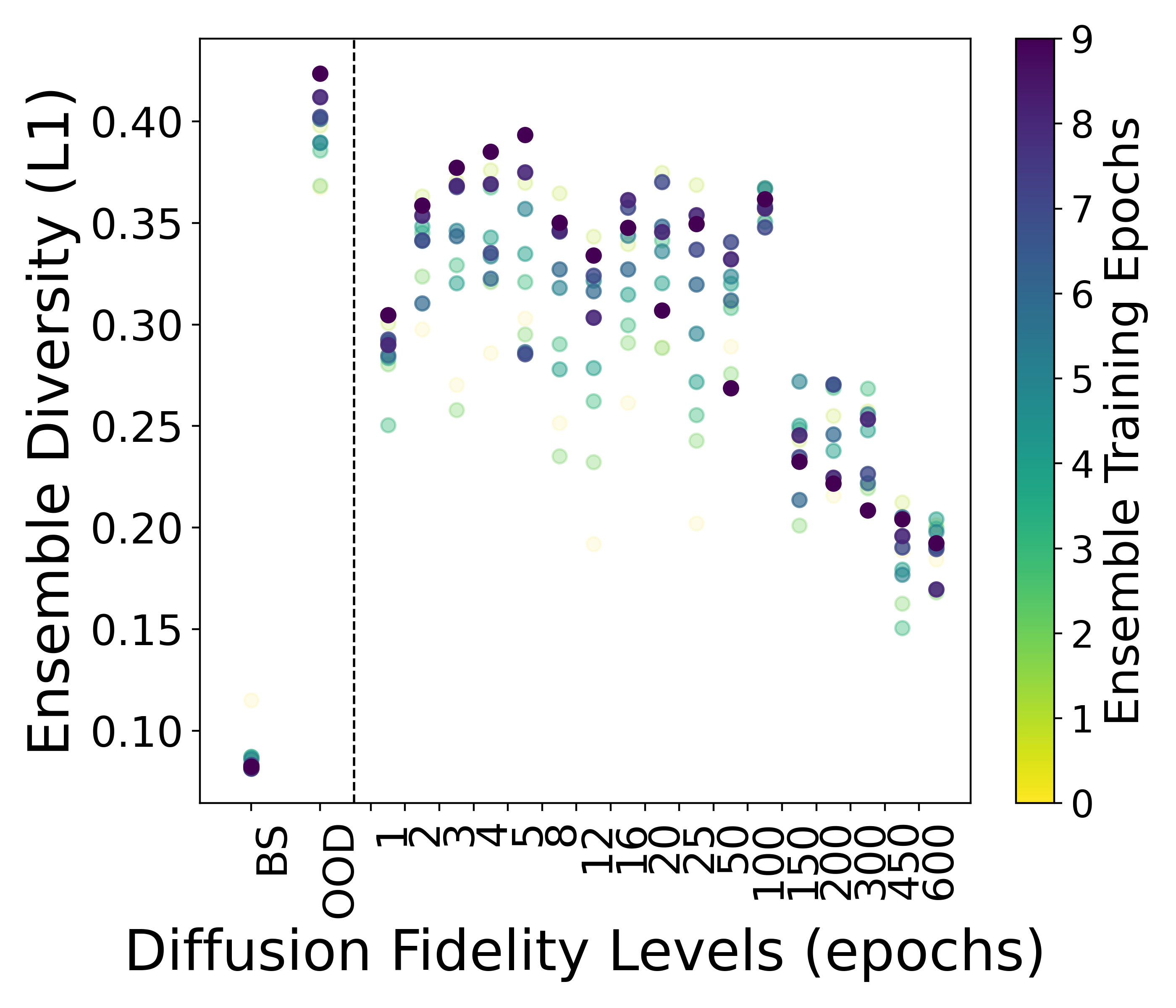}
        \caption*{L1} 
    \end{subfigure}
    \begin{subfigure}[b]{.195\textwidth}
        \centering
        \includegraphics[width=\linewidth]{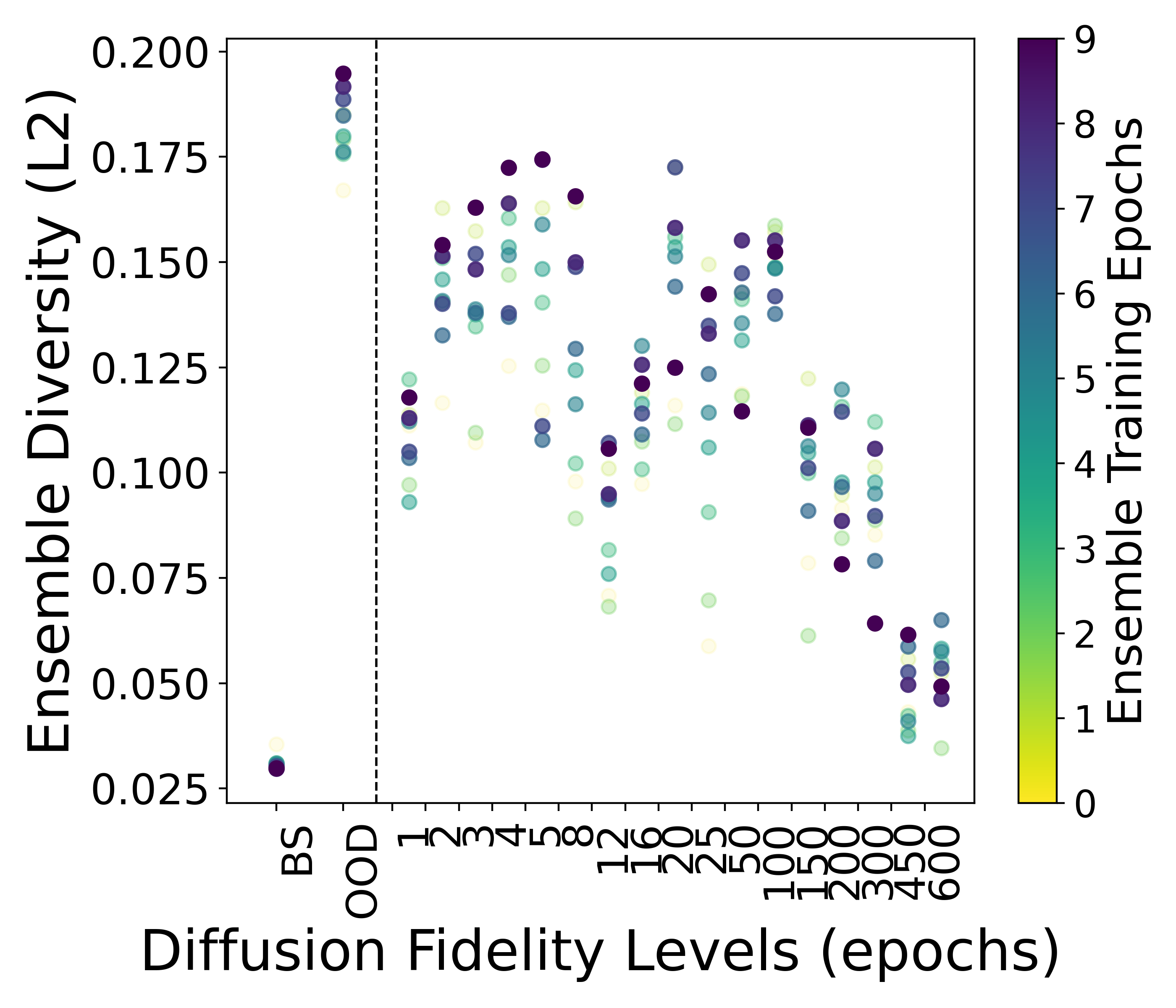}
        \caption*{L2} 
    \end{subfigure}
    \vspace{-5pt}
    \caption{ColorDSprites} 
\end{subfigure}
\begin{subfigure}[b]{\textwidth}
    \vspace{10pt}
    \centering  
    
    \begin{subfigure}[b]{.195\columnwidth}
        \includegraphics[width=\linewidth]{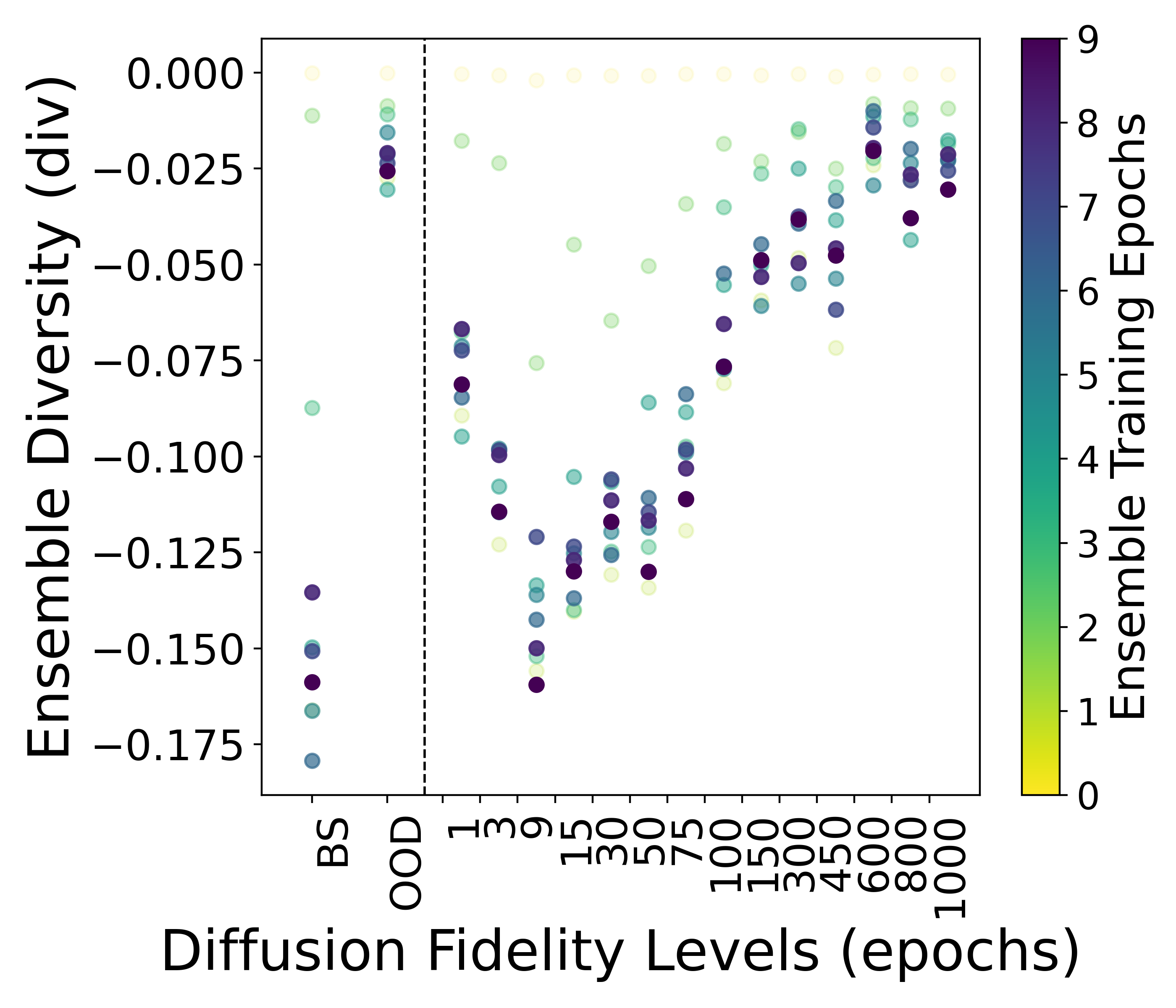}
        \caption{UTKFace} 
    \end{subfigure}
    \begin{subfigure}[b]{.195\textwidth}
        \centering
        \includegraphics[width=\linewidth]{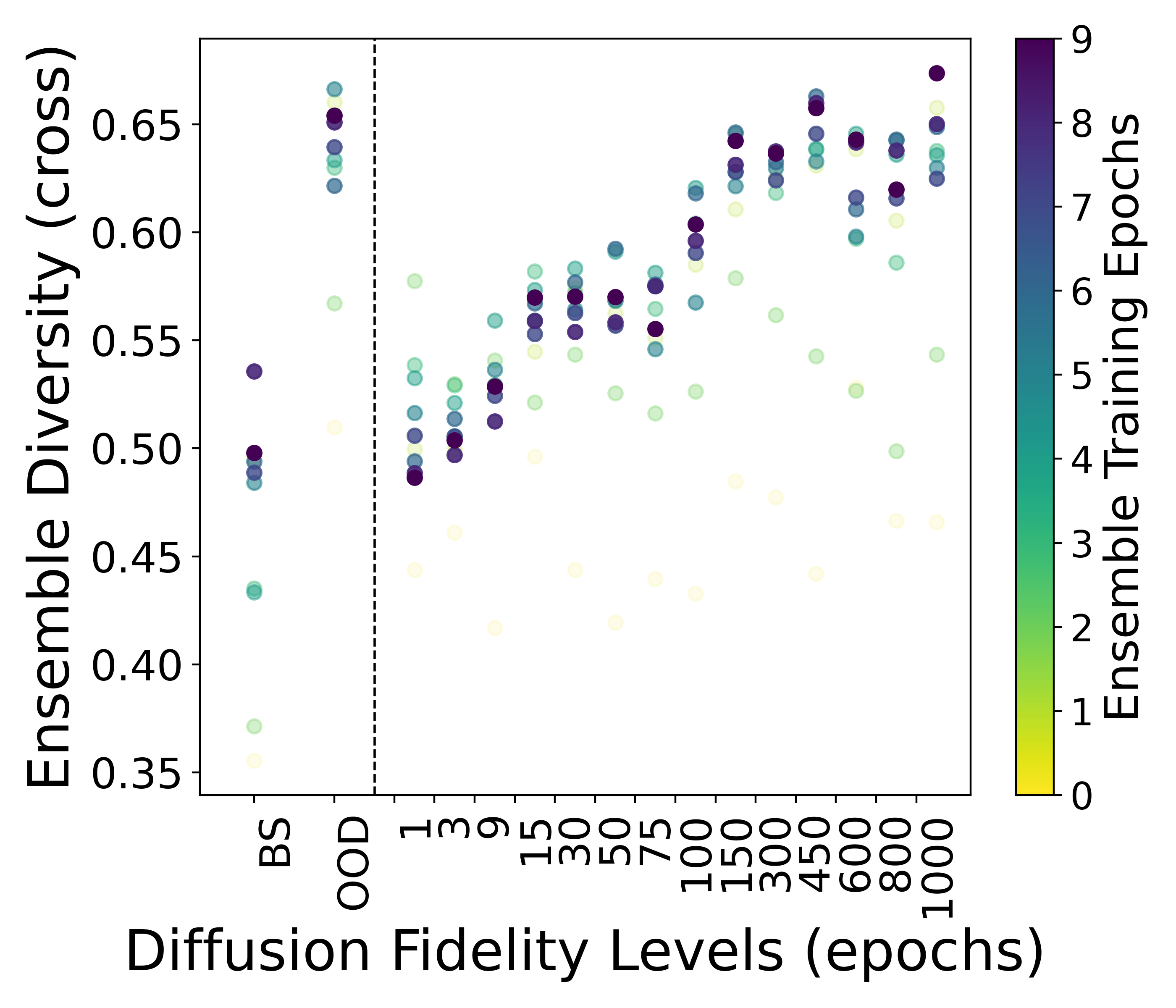}
        \caption*{cross} 
    \end{subfigure}
    \begin{subfigure}[b]{.195\textwidth}
        \centering
        \includegraphics[width=\linewidth]{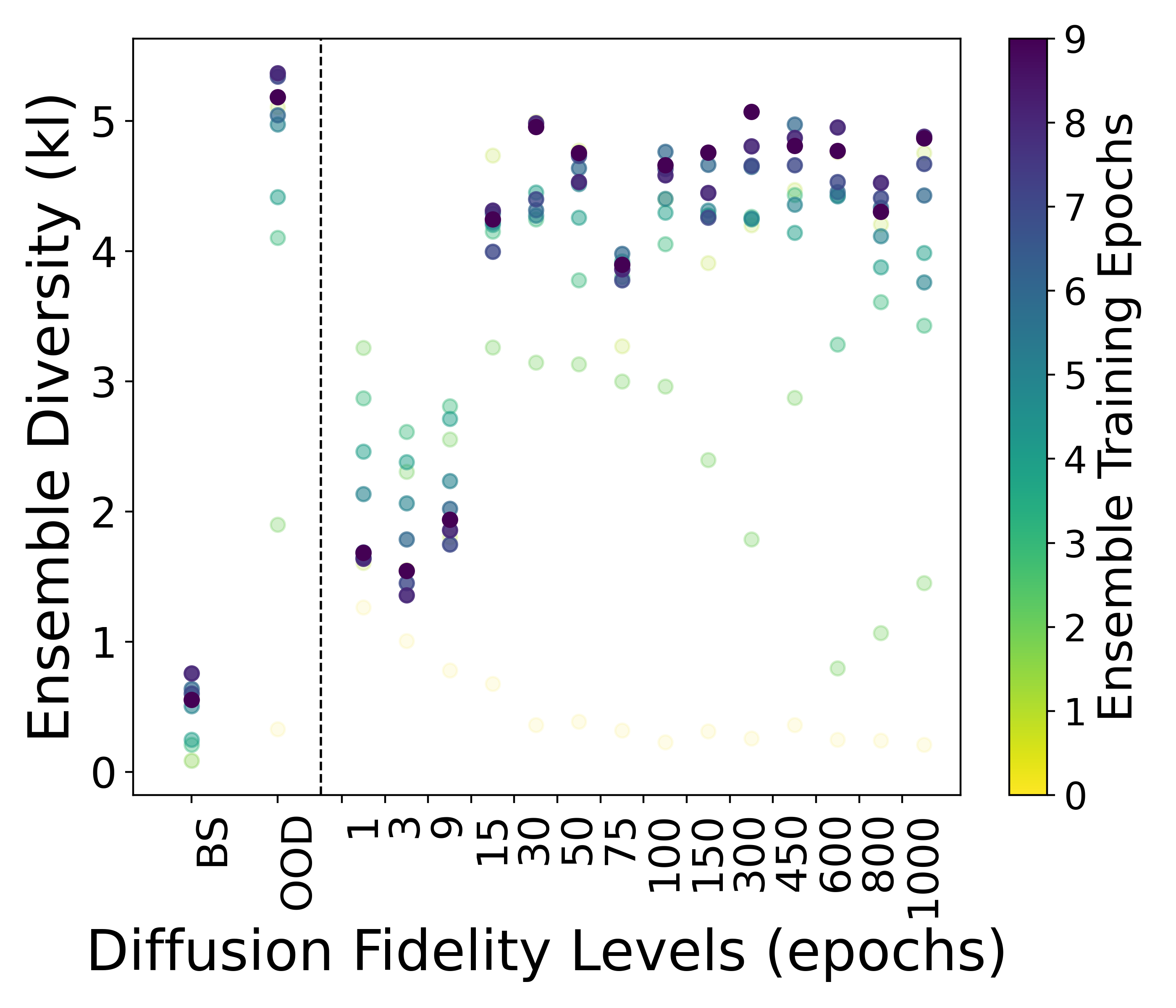}
        \caption*{kl} 
    \end{subfigure}
    \begin{subfigure}[b]{.195\textwidth}
        \centering
        \includegraphics[width=\linewidth]{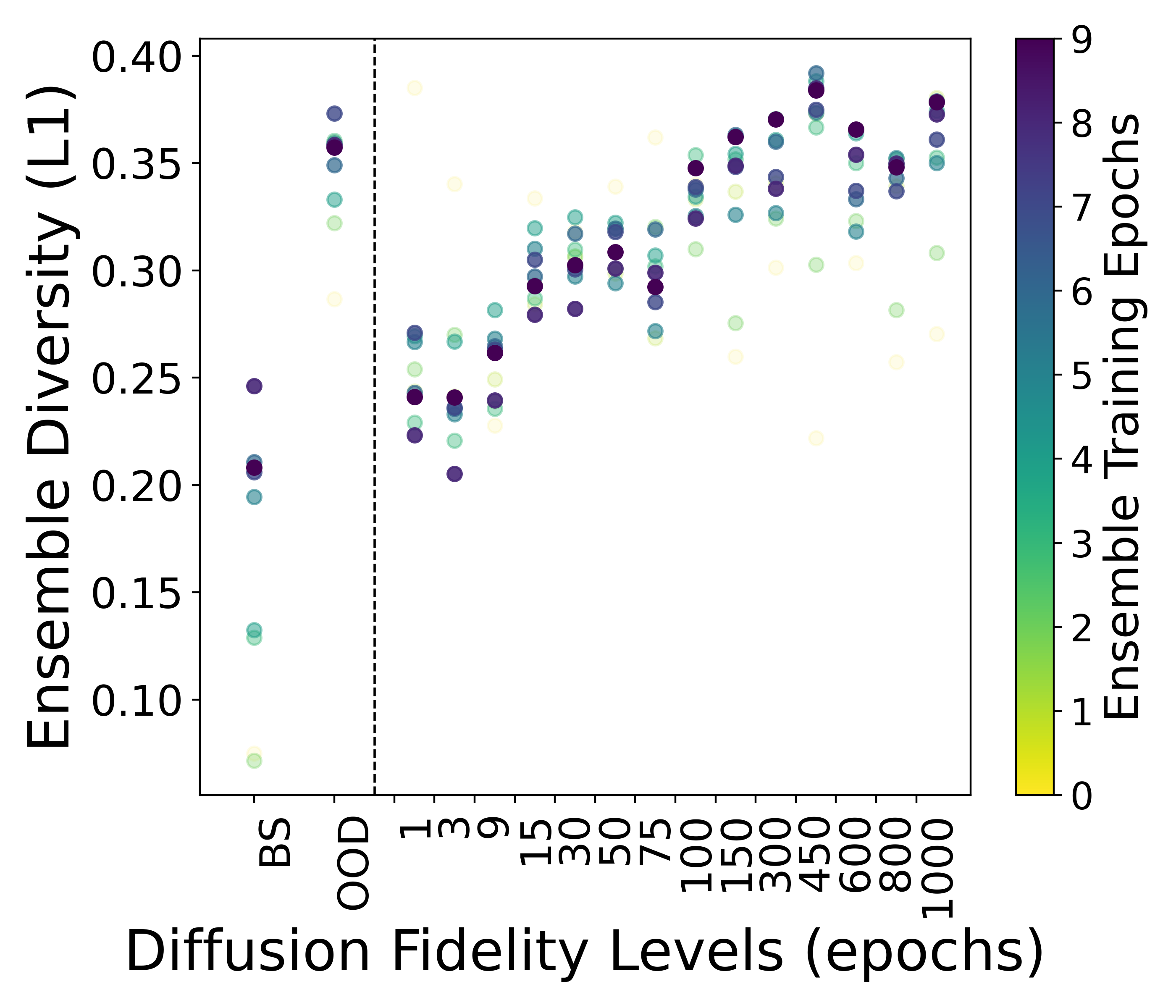}
        \caption*{L1} 
    \end{subfigure}
    \begin{subfigure}[b]{.195\textwidth}
        \centering
        \includegraphics[width=\linewidth]{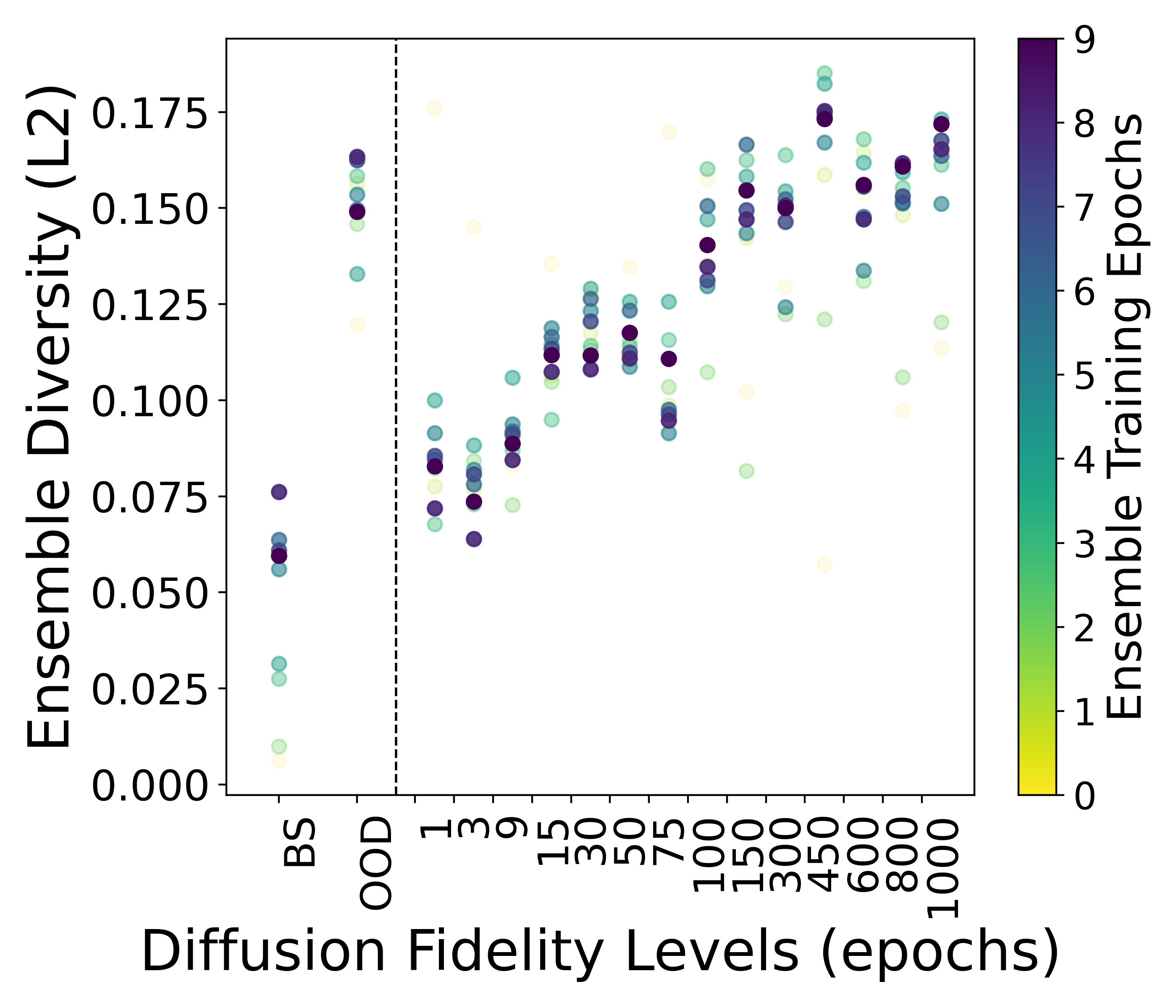}
        \caption*{L2} 
    \end{subfigure}
    \vspace{-5pt}
    \caption{UTKFace} 
\end{subfigure}
\begin{subfigure}[b]{\textwidth}
    \vspace{10pt}
    \centering  
    
    \begin{subfigure}[b]{.195\columnwidth}
        \includegraphics[width=\linewidth]{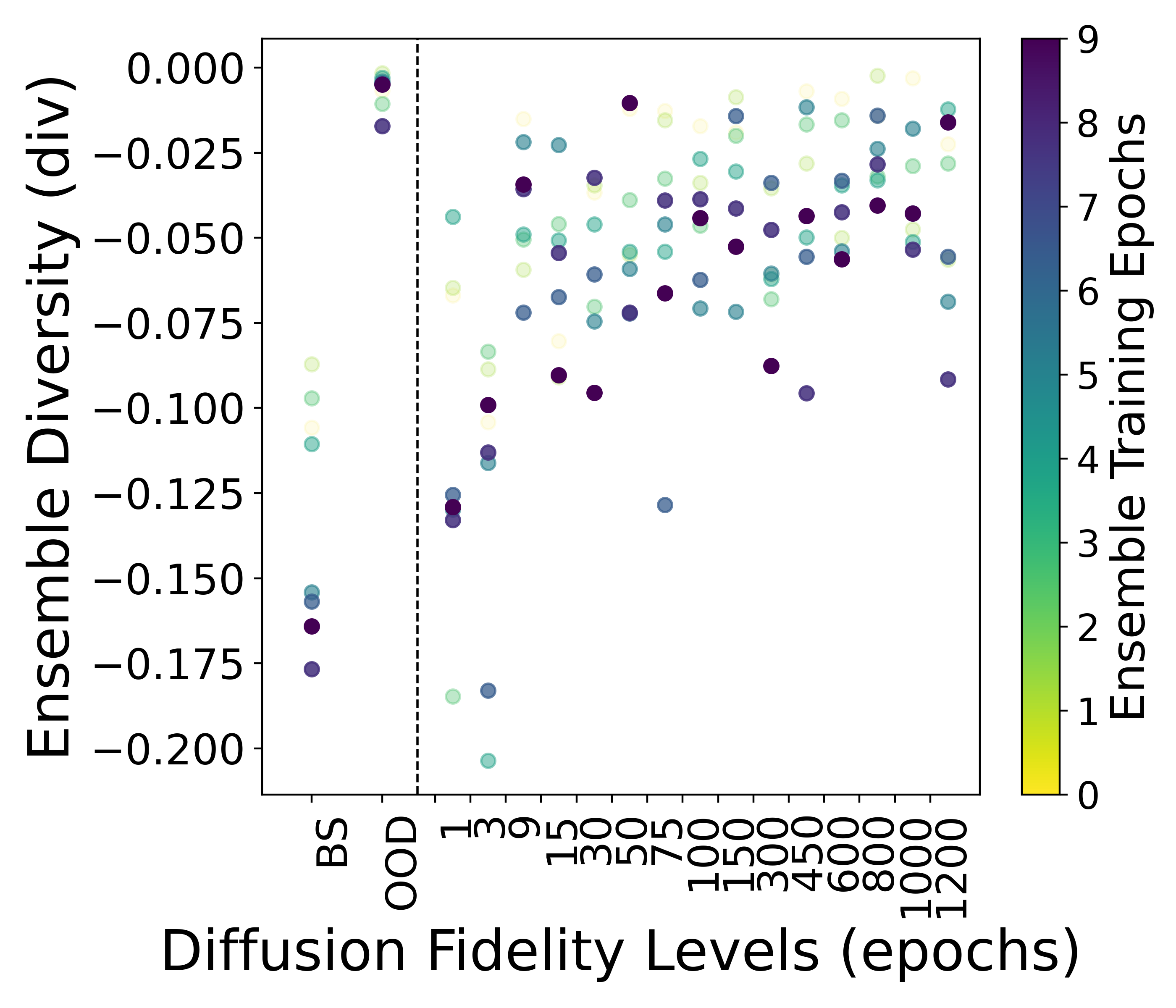}
        \caption{CelebA} 
    \end{subfigure}
    \begin{subfigure}[b]{.195\textwidth}
        \centering
        \includegraphics[width=\linewidth]{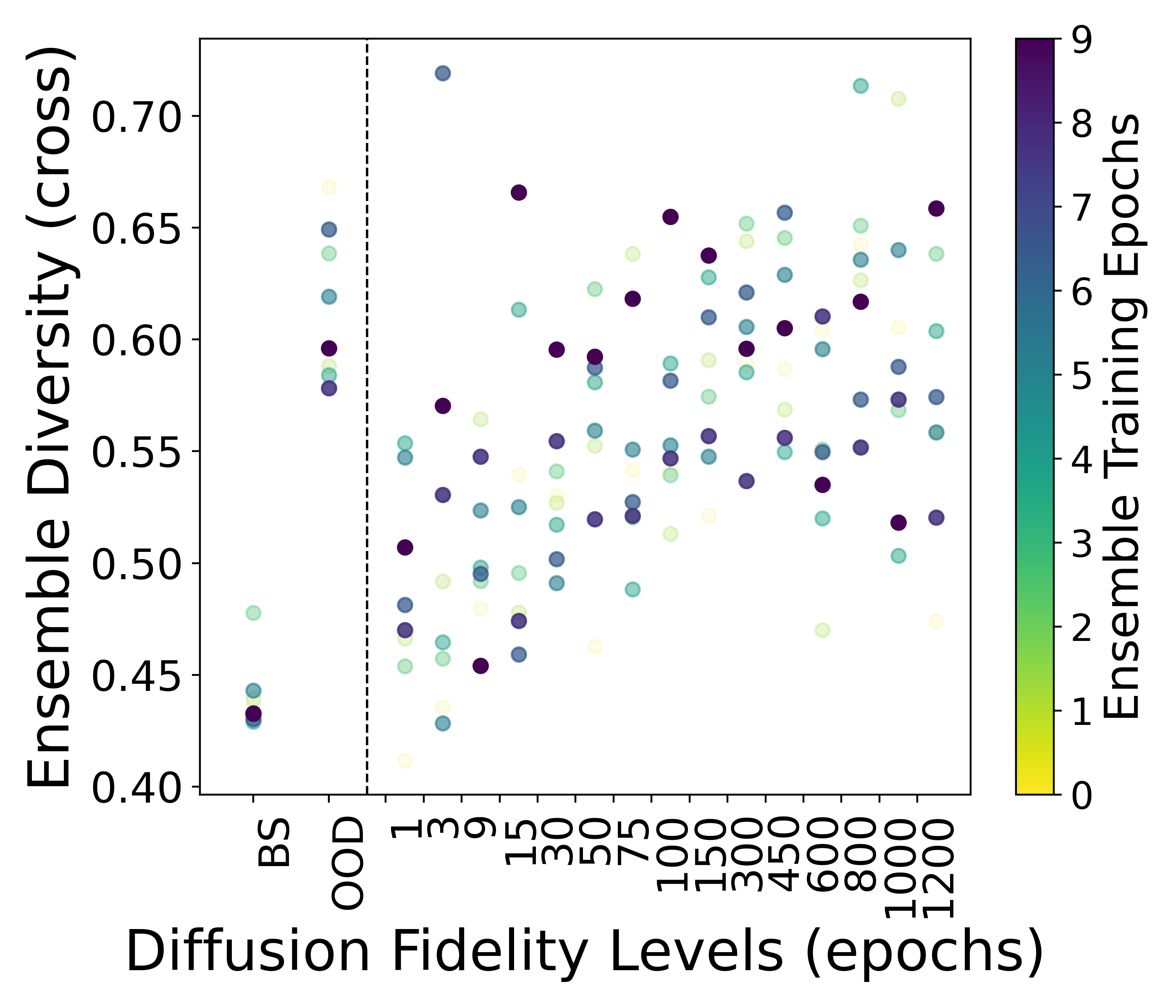}
        \caption*{cross} 
    \end{subfigure}
    \begin{subfigure}[b]{.195\textwidth}
        \centering
        \includegraphics[width=\linewidth]{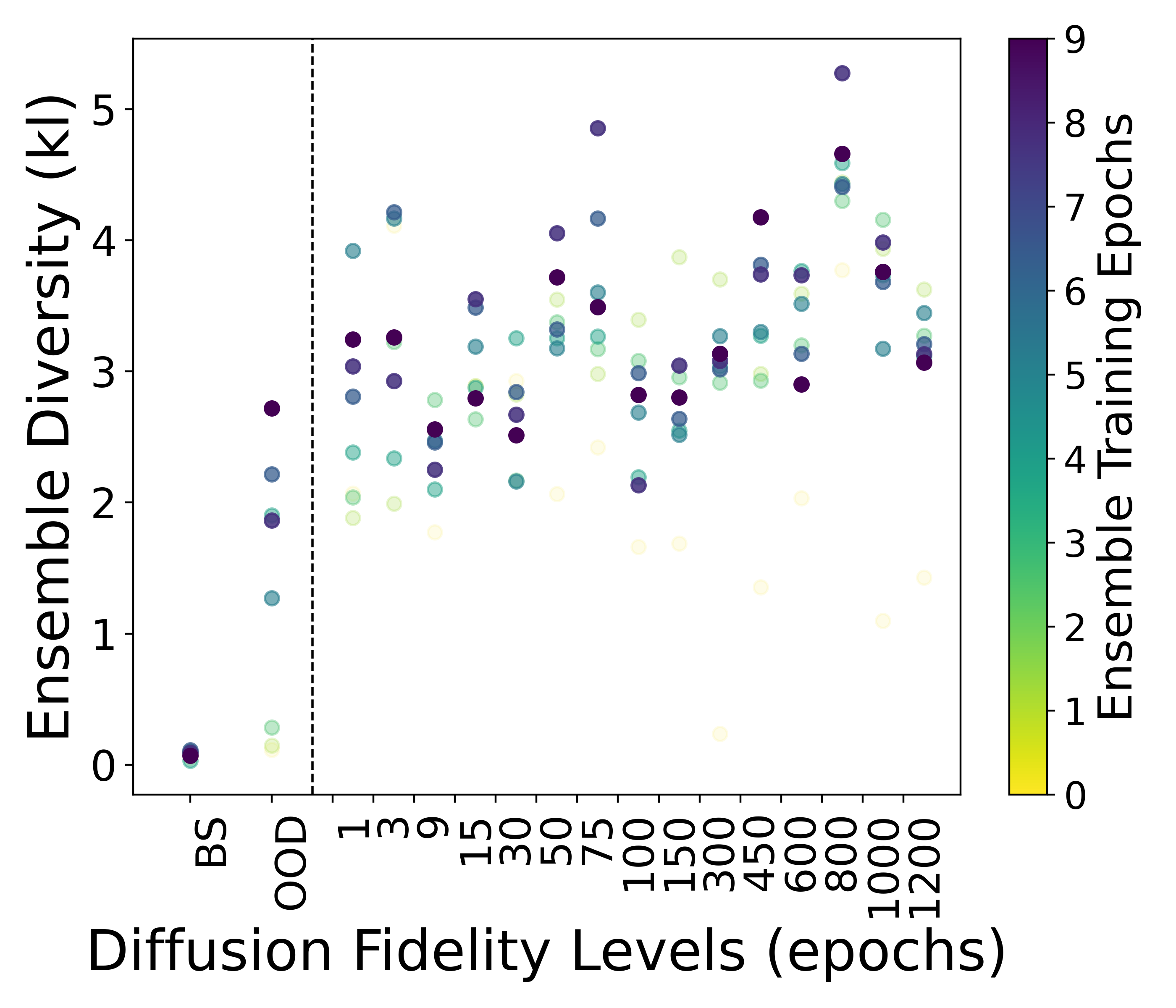}
        \caption*{kl} 
    \end{subfigure}
    \begin{subfigure}[b]{.195\textwidth}
        \centering
        \includegraphics[width=\linewidth]{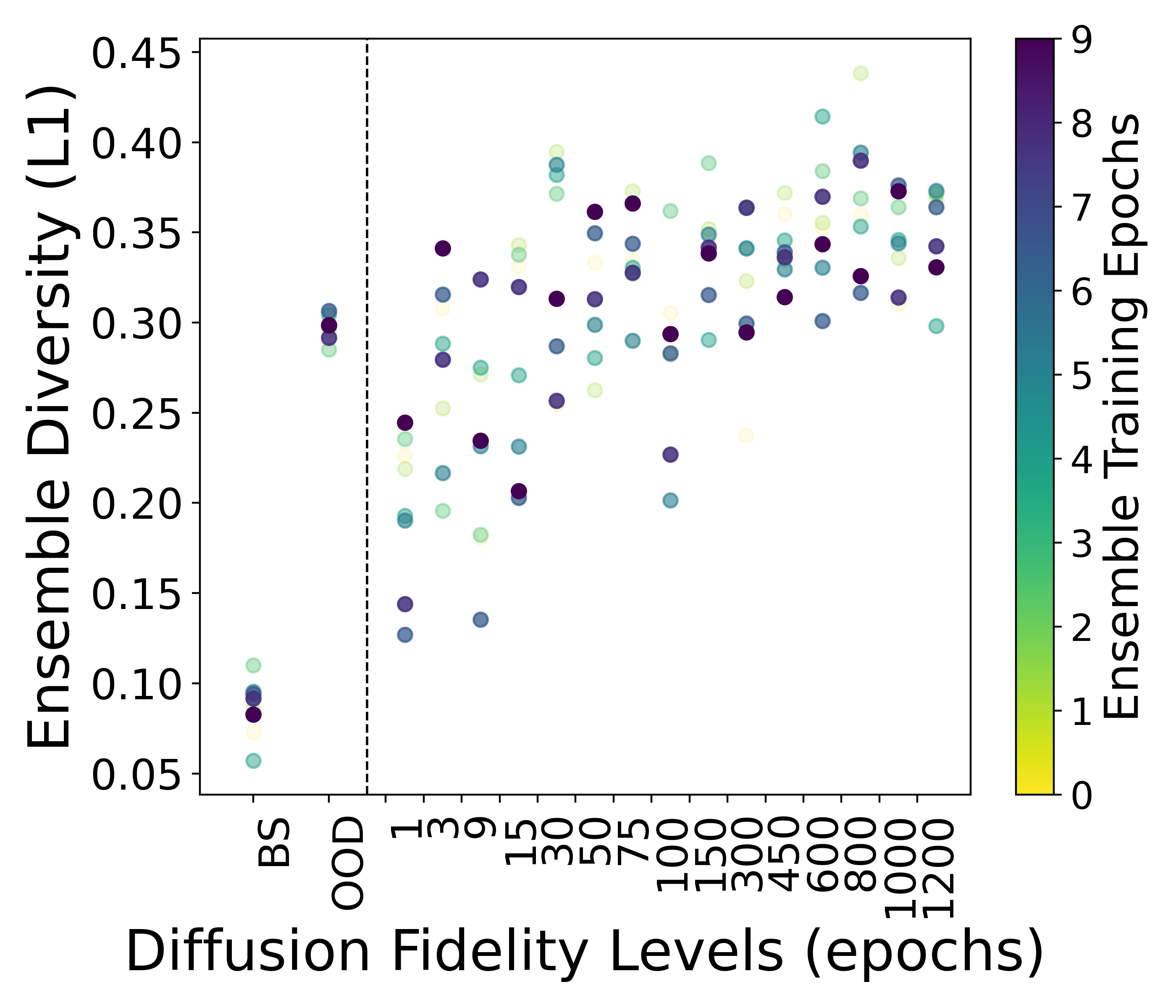}
        \caption*{L1} 
    \end{subfigure}
    \begin{subfigure}[b]{.195\textwidth}
        \centering
        \includegraphics[width=\linewidth]{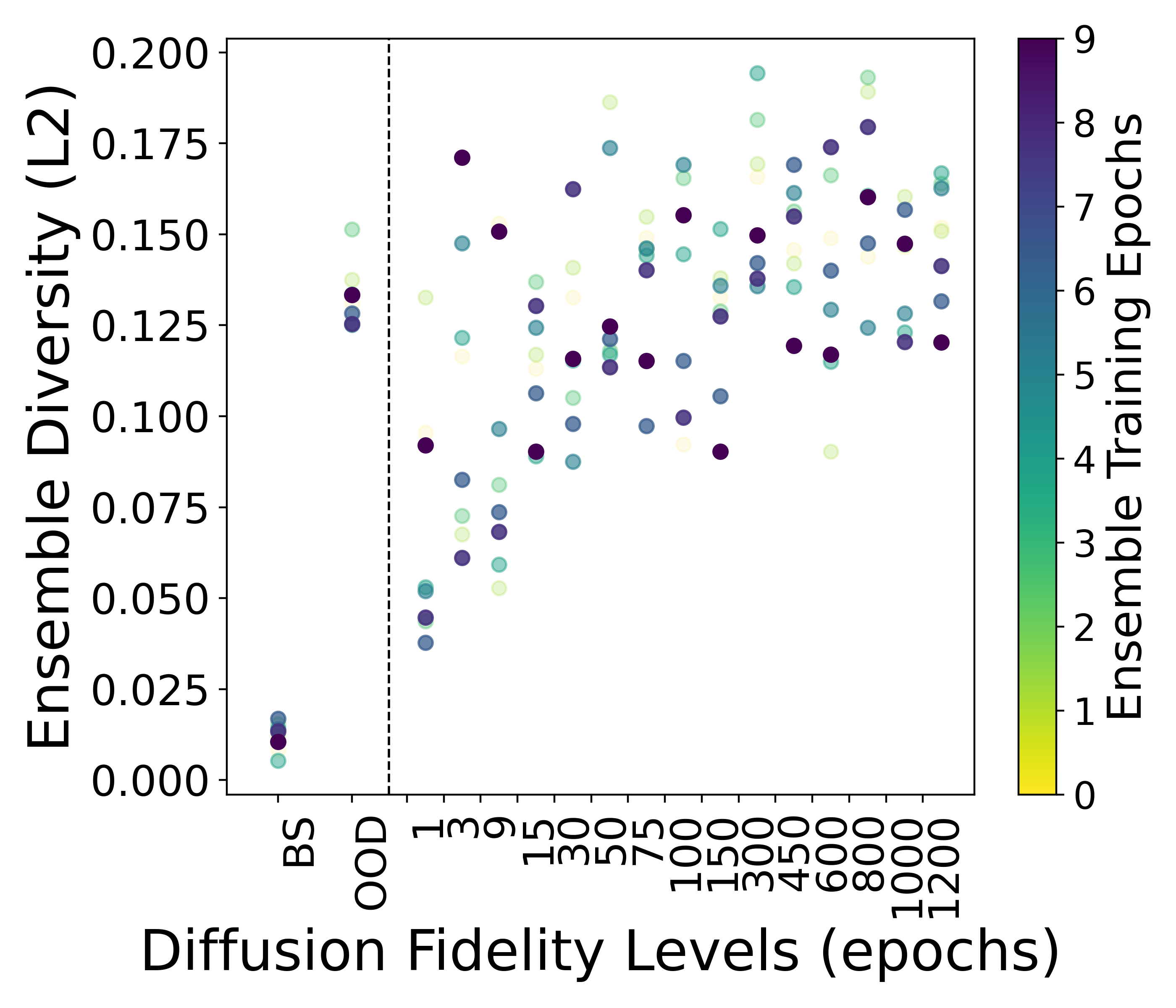}
        \caption*{L2} 
    \end{subfigure}
    \vspace{-5pt}
    \caption{CelebA} 
\end{subfigure}
\caption{Ensemble diversity as enforced via samples from diffusion models trained a different fidelity levels (i.e. diffusion training epochs).} 
\label{sup:fig:fidelity_vs_diversity}
% \vspace{-10pt}
\end{figure*}

We perform experiments whereby an ensemble of 100 ResNet-18 models is trained separately with respect to all diversification objectives considered. Figure \ref{sup:fig:fidelity_vs_diversity} shows the diversification results as a function of the fidelity of the DPM used to sample the diversification set. Although we observe that the diversification level obtained is dependent on the diffusion fidelity level, suggesting the need for appropriate early stopping procedures to achieve increased ensemble prediction diversity, we find DiffDiv to not be overly sensitive to this choice. In fact, we observe broad areas with similar diversity levels across several DPM fidelities.  While the trends vary mildly across different disagreement objectives, we find the diversification maximized in ColorDSprites and UTKFace coherently with the analysis in Section \ref{sec:res:diff_disentanglement}, where improved diversification is achieved around the \emph{originative} interval. For ColorDSprites, this is realized in around 20 DPM training epochs, for UTKFace, it is realized in around 800 DPM training epochs, while for CelebA it is around 1000 DPM training epochs. 

\begin{figure}[t]
\centering
\includegraphics[width=.98\linewidth]{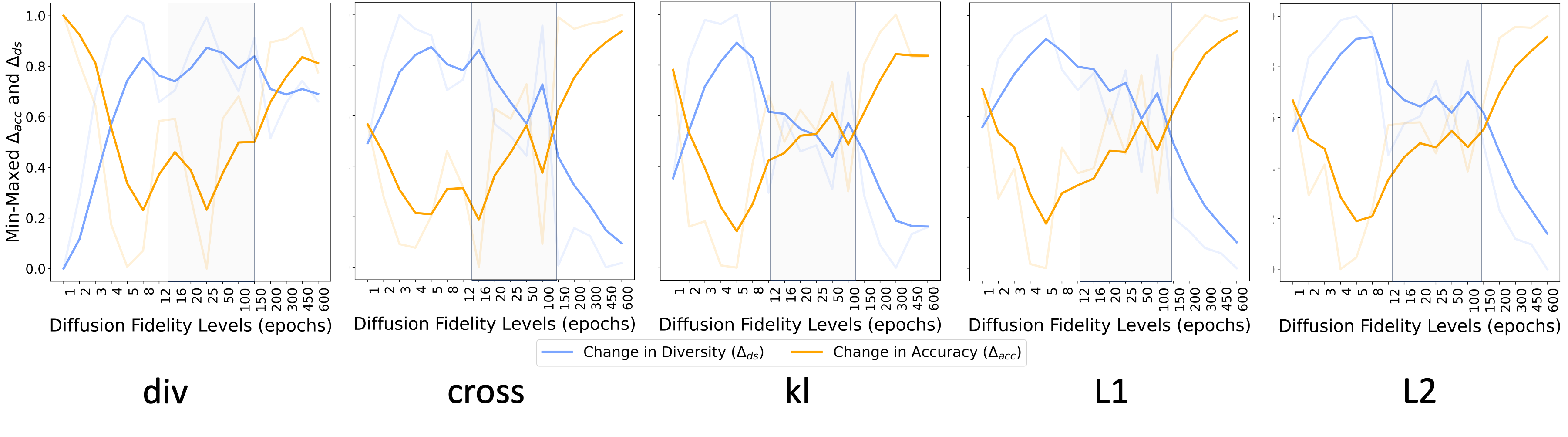}
\caption{Min-Maxed change in accuracy and diversity by ensembles trained with diffusion-augmented samples on ColorDSprites, with respect to all considered diversification methods. The \emph{originative} stage, as qualitatively identified in the experiments (see \autoref{fig:fidelity_sampling} and \autoref{fig:fidelity_vs_diversity}) is shown in gray. The areas primarily highest in diversity, with the least change in accuracy, are significant of the originative stage. We mark with red discontinuous vertical bars the stage used for the experiments } 
\label{fig:originative_search} 
\vspace{-5pt}
\end{figure}

Interestingly, our results suggest how ensemble diversification metrics can be a viable proxy for appropriate \emph{originative} DPM training. In \autoref{fig:originative_search}, we observe the min-maxed change in accuracy and change in diversity by the ensembles with respect to baseline training on ColorDSprites. In the figure, we observe similar trends across all diversification methods, whereby the \emph{originative} interval (highlighted in gray) is primarily identified by the highest changes in diversity, while also considering the least drop in accuracy. These results confirm our previous supervised findings (\autoref{fig:fidelity_sampling}). Imporantly, we find that only when jointly looking at both diversity and validation performance we can best identify the relevant areas around the generative stage. Under this light, ensemble performance/diversity validation metrics can be directly leveraged for DPM early stopping logic. Our results align with our previous observations, where the highest number of ood-generated samples to lie within the DPM training intervals achieve the highest change in diversity while maintaining good classification performance (\autoref{fig:accuracy_diversity_deltas}.)

\begin{table}[t]
  \caption{Disagreement $\gamma$ used in our experiments.}  % Replace with your actual title
  \label{suppl:tab:dis_hyper}      % Replace with your actual label
  \centering
  \begin{tabular}{lccccc}
    \toprule
    & \textbf{L2} & \textbf{L1} & \textbf{cross} & \textbf{div} & \textbf{kl} \\
    \midrule
    \textbf{ColorDSPrites} & 5.0 & 1.5 & 0.5 & 0.5 & 0.1 \\
    \textbf{UTKFace}  & 5.0 & 2.0 & 2.0 & 5.0 & 0.2 \\
    \textbf{CelebA}  & 5.0 & 2.0 & 2.0 & 5.0 & 0.2 \\
    \bottomrule
  \end{tabular}
\end{table}

\begin{table}
\caption{Comparison between the ensemble accuracy before (Ens) and after model selection (Select). Each subset of models in the ensemble is chosen dynamically while keeping all models with cue averted tendencies in \autoref{tab:diversity:all}}
\centering
\resizebox{1\linewidth}{!}{

\begin{tabular}{@{}lccccccccc}\toprule
Dataset $\rightarrow$ & \multicolumn{2}{c}{ColorDSprites} & \multicolumn{2}{c}{UTKFace} & \multicolumn{2}{c}{CelebA} \\\cmidrule(lr){2-3}\cmidrule(lr){4-5}\cmidrule(lr){6-7}
Obj. $\downarrow$ & Valid. Acc. (Ens) & Valid. Acc. (Select) & Valid. Acc. (Ens) & Valid. Acc. (Select) & Valid. Acc. (Ens) & Valid. Acc. (Select) \\\midrule

Baseline & $1.000 \pm 0.00$ & $1.000 \pm 0.00$ & $0.920 \pm 0.02$ & $0.943 \pm 0.00$ & $0.857 \pm 0.01$ & $0.873 \pm 0.00$ \\
Cross    & $0.856 \pm 0.16$ & $0.945 \pm 0.14$ & $0.836 \pm 0.05$ & $0.856 \pm 0.07$ & $0.745 \pm 0.10$ & $0.828 \pm 0.06$ \\
Div      & $0.916 \pm 0.13$ & $0.980 \pm 0.07$ & $0.826 \pm 0.05$ & $0.868 \pm 0.06$ & $0.857 \pm 0.01$ & $0.873 \pm 0.00$ \\
KL       & $0.786 \pm 0.20$ & $0.872 \pm 0.22$ & $0.837 \pm 0.06$ & $0.858 \pm 0.08$ & $0.672 \pm 0.07$ & $0.713 \pm 0.09$ \\
L1      & $0.784 \pm 0.20$ & $0.861 \pm 0.23$ & $0.816 \pm 0.11$ & $0.824 \pm 0.11$ & $0.659 \pm 0.12$ & $0.737 \pm 0.12$ \\
L2      & $0.762 \pm 0.22$ & $0.864 \pm 0.26$ & $0.757 \pm 0.12$ & $0.776 \pm 0.13$ & $0.650 \pm 0.11$ & $0.716 \pm 0.12$ \\\bottomrule
\end{tabular}
\label{tab:model_selection}
}
\end{table}
\subsection{Diversification Leads to Ensemble Models Attending to Different Cues} \label{sup:sec:res_diversification_comparison}
\begin{figure}[t]
    \begin{subfigure}[b]{\linewidth}
        \centering
        \begin{subfigure}[b]{\linewidth}
            \includegraphics[trim={0 0 0 50}, clip, width=\linewidth]{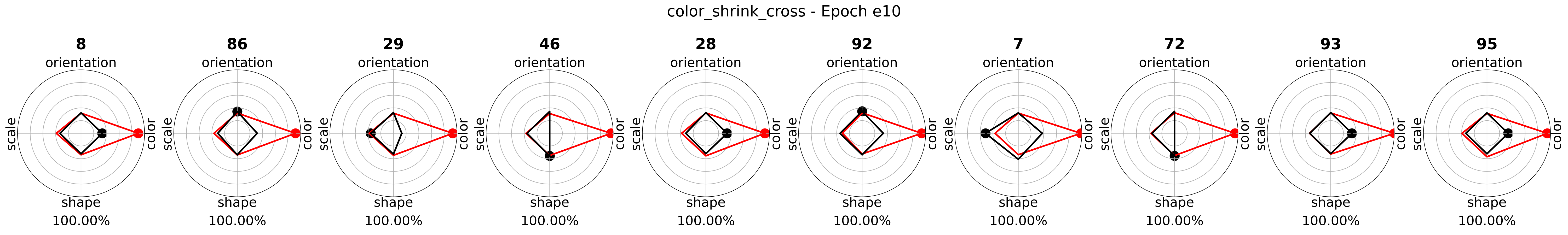}
            \caption*{ColorDSprites}
        \end{subfigure}%
        \hfill
        \begin{subfigure}[b]{.93\linewidth}
            \includegraphics[trim={0 0 0 50}, clip, width=\linewidth]{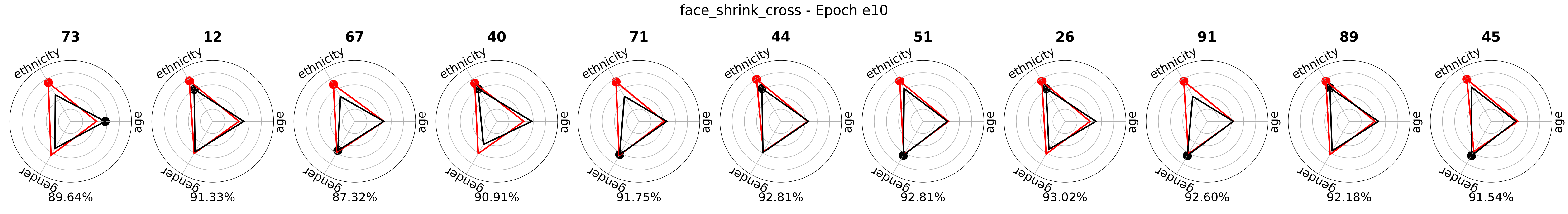}
            \caption*{UTKFace}
        \end{subfigure}
        \hfill
        \begin{subfigure}[b]{.93\linewidth}
            \includegraphics[trim={0 0 0 50}, clip, width=\linewidth]{figures/model_comparison/CelebA_shrink_std_e8_model_comparison.png}
            \caption*{CelebA}
        \end{subfigure}
    \vspace{5pt}
    \caption{\emph{ood} Disagreement}
    \vspace{10pt}
    \label{fig:diversity_comparison:ood}
    \end{subfigure}
    \begin{subfigure}[b]{\linewidth}
        \centering
        \begin{subfigure}[b]{\linewidth}
            \includegraphics[trim={0 0 0 50}, clip, width=\linewidth]{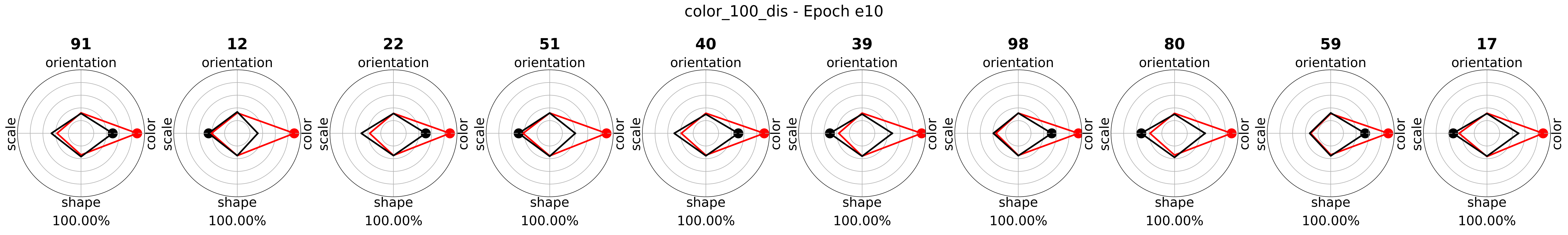}
            \caption*{ColorDSprites}
        \end{subfigure}%
        \hfill
        \begin{subfigure}[b]{.93\linewidth}
            \includegraphics[trim={0 0 0 50}, clip, width=\linewidth]{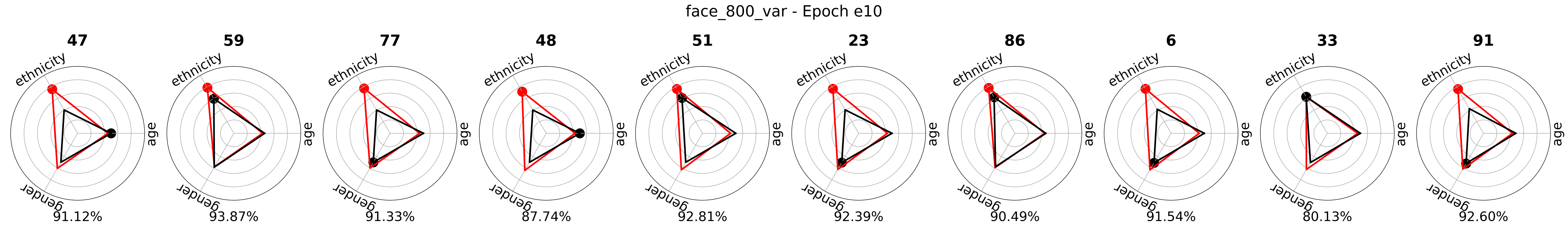}
            \caption*{UTKFace}
        \end{subfigure}
        \hfill
        \begin{subfigure}[b]{.93\linewidth}
            \includegraphics[trim={0 0 0 50}, clip, width=\linewidth]{figures/model_comparison/CelebA_1200_std_e8_model_comparison.png}
            \caption*{CelebA}
        \end{subfigure}
    \caption{Diffusion Disagreement}
    \vspace{5pt}
    \label{fig:diversity_comparison:diff}
    \vspace{10pt}
    \end{subfigure}
    \caption{Comparison of 10 diversified models when training the ensemble while using (a) feature-uncorrelated \emph{ood} data and (b) Diffusion Samples.}
    \label{fig:diversity_comparison}
\end{figure}

Figure \ref{fig:diversity_comparison} illustrates a feature-centric description of 10 ensemble models trained with a diversification objective on \emph{ood} data (a) and Diffusion generated counterfactuals (b). The variation across models is evident: several models substantially reduce their dependency on the leading cues of the respective datasets (black edges), diverging considerably from the almost identical configurations present in the baseline ensemble (red edges). For some of the models, in fact, the averted attention on the main shortcut cue leads to increased reliance on one of the other observed features (e.g. scale and age for models 7 (ColorDsprites), 73 (UTKFace) and 5 (CelebA) in \autoref{fig:diversity_comparison:ood}, and models 80 (ColorDsprites), 47 (UTKFace) and 61 (CelebA)) in \autoref{fig:diversity_comparison:diff}).

\subsection{On the Influence of an Increased Number of \emph{ood} Samples for Ensemble Disagreement} \label{sup:sec:res_diversification_all_ood}

As per the original objective, with DPM sampling we aim to circumvent the diversification dependency on Out-Of-Distribution data, which is often not readily accessible and can be costly to procure. We test this dependency further and assess the quality of the diversification results when matching the number of \emph{ood} data used for diversification to the original training data for the ensemble. We report in \autoref{tbl:suppl_ood_full} our findings. We observe the quality of the disagreement on ColorDSprites to only marginally benefit from additional disagreement samples, with approximately $1\%$ to $7\%$ more of the models to avert their attention from the shortcut cue \emph{color} as compared to the original experiments. On the other hand, we observe a strong improvement in the diversification for UTKFace, mainly registered via the \emph{div} objective, where $24\%$ of the models averted their attention from the \emph{ethnicity} shortcut, as opposed to the original $6\%$ in our previous experiments,  while maintaining high predictive performance on the validation set. We observe marginal improvements on the other objectives, with approximately $4\%$ to $8\%$ additional models achieving cue aversion. We speculate this gain to be due to the higher complexity of the features within the data, which may require additional specimens for appropriate diversification.

\begin{table}[t] % table* will make the table span both columns
\centering
\caption{Diversification results on ColorDSprites, UTKFace, and CelebA when using the same number of \emph{ood} samples as the training dataset. The feature columns report the fraction of models (in each row) biased towards the feature. The final column reports the average validation accuracy for the ensemble when tested on a left-out feature-correlated \emph{diagonal} set, of the same distribution as the original training data}

\resizebox{1\linewidth}{!}{
\begin{tabular}{@{}lccccccccccccc}\toprule
Dataset $\rightarrow$ & \multicolumn{5}{c}{ColorDSprites} & \multicolumn{4}{c}{UTKFace} & \multicolumn{3}{c}{CelebA} \\\cmidrule(lr){2-6}\cmidrule(lr){7-10}\cmidrule(lr){11-13} 
Obj. $\downarrow$ & Color ($\downarrow$)  & Orient.  & Scale  & Shape  & Acc. ($\uparrow$)   
        & Age  & Ethnicity ($\downarrow$)  & Gender  & Acc. ($\uparrow$)  
        & Oval Face & Pale Skin ($\downarrow$) & Acc. ($\uparrow$)\\\toprule

Baseline & $1.00$          & $0.00$                & $0.00$          & $0.00$          & $1.000$\std{$0.00$}   
         & $0.00$          & $1.00$              & $0.00$             & $0.920$\std{$0.02$}  
         & $0.00$          & $1.00$            & $0.857$\std{$0.01$} \\

Cross    & $0.92$          & $0.00$                & $0.08$          & $0.00$          & $0.849$\std{$0.16$}   
         & $0.00$          & $0.73$              & $0.27$             & $0.858$\std{$0.04$}  
         & $0.01$          & $0.99$            & $0.751$\std{$0.07$} \\

Div      & $0.82$          & $0.00$                & $0.18$          & $0.00$          & $0.820$\std{$0.19$}   
         & $0.00$          & $0.76$              & $0.24$             & $0.844$\std{$0.04$}  
         & $0.00$          & $1.00$            & $0.843$\std{$0.03$} \\

KL       & $0.90$          & $0.03$                & $0.05$          & $0.02$          & $0.801$\std{$0.20$}   
         & $0.02$          & $0.68$              & $0.30$             & $0.820$\std{$0.07$}  
         & $0.00$          & $1.00$            & $0.812$\std{$0.04$} \\

L1       & $0.90$          & $0.01$                & $0.07$          & $0.02$          & $0.799$\std{$0.21$}   
         & $0.00$          & $0.66$              & $0.34$             & $0.832$\std{$0.09$}  
         & $0.14$          & $0.86$            & $0.724$\std{$0.11$} \\

L2       & $0.86$          & $0.04$                & $0.08$          & $0.02$          & $0.745$\std{$0.22$}   
         & $0.08$          & $0.58$              & $0.34$             & $0.761$\std{$0.15$}  
         & $0.12$          & $0.88$            & $0.651$\std{$0.10$} \\\bottomrule
\end{tabular}
\label{tbl:suppl_ood_full}
}

\end{table}

\begin{figure}[t]
\vspace{-15pt}
\centering
\includegraphics[width=\linewidth]{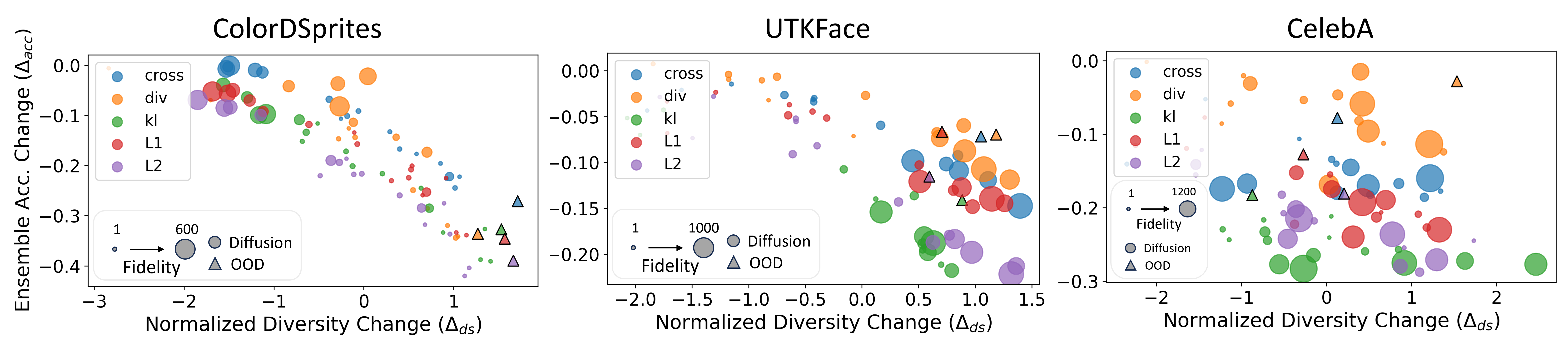} 
\caption{The relationship between the change in normalized classification prediction diversity ($\Delta_{ds}$) and the change in validation accuracy of the ensemble ($\Delta_{acc}$), when trained with samples from DPMs at varying levels of fidelities. The $\Delta$s are computed with respect to baseline ensemble training, with no diversification objective. We also compare the metrics achieved by diversification with non-correlated, off-diagonal, \emph{ood} data from the respective datasets. } 
\label{fig:accuracy_diversity_deltas}
\vspace{-10pt}

\end{figure}

\paragraph{Increased Ensemble Disagreement Negatively Correlates with Ensemble \emph{id} Performance:} 
Achieving ensemble diversity by disagreement on unlabelled \emph{ood} samples has previously been shown to negatively impact \emph{id} ensemble performance~\citep{pagliardini2022agree} (suppl. \autoref{fig:calibration}), while improving \emph{ood} performance on downstream tasks associated with non-shortcut cues. Generally, ensemble model selection has been an effective method to prune models that misalign with the original classification objective~\citep{lee2022diversify}.
In this section, we wish to understand this relationship under the scope of Diffusion-guided diversification.
In \autoref{fig:accuracy_diversity_deltas} we observe the change in ensemble average accuracy $\Delta_{acc}$ as a function of the change in diversity $\Delta_{ds}$, spanning all the diversification objectives. We highlight a few important observations. First, increased ensemble diversity via disagreement negatively correlates with average ensemble accuracy, without additional model selection mechanisms. Second, real \emph{ood} data achieves the highest diversity gain, comparable to the diffusion samples at the \emph{originative} stage (\S \ref{sec:res:diff_disentanglement}), $\approx [10, 100]$ for ColorDSprites, $\approx [450, 800]$ for UTKFace, and $\approx [1000, 1200]$ for CelebA. Lastly, the diversification objectives show comparable trends, with the $div$ objective displaying marginally better diversification/accuracy performance than the others on both tasks. 

\end{document}